\documentclass[manuscript]{acmart}
\usepackage{multirow}
\usepackage{bibunits}
\usepackage{tikz}
\usetikzlibrary{arrows.meta,shapes.geometric}
\usepackage{tabularx}
\usepackage{placeins}
\usepackage{needspace}
\usepackage{tcolorbox}
\tcbuselibrary{breakable,skins}
\usepackage{fvextra}
\newif\ifrevisionmarks
\revisionmarksfalse
\definecolor{RevisionBlue}{HTML}{174D91}
\newcommand{\rev}[1]{{\ifrevisionmarks\color{RevisionBlue}\fi #1}}
\newenvironment{revision}{\par\begingroup\ifrevisionmarks\color{RevisionBlue}\fi}{\par\endgroup}
\definecolor{RevisionTeal}{HTML}{006B5E}
\newcommand{\revM}[1]{{\ifrevisionmarks\color{RevisionTeal}\fi #1}}
\newenvironment{revisionM}{\par\begingroup\ifrevisionmarks\color{RevisionTeal}\fi}{\par\endgroup}
\definecolor{RevisionPurple}{HTML}{7030A0}
\newcommand{\revN}[1]{{\ifrevisionmarks\color{RevisionPurple}\fi #1}}
\newenvironment{revisionN}{\par\begingroup\ifrevisionmarks\color{RevisionPurple}\fi}{\par\endgroup}
\definecolor{RevisionOrange}{HTML}{A34B00}
\newcommand{\revP}[1]{{\ifrevisionmarks\color{RevisionOrange}\fi #1}}
\newenvironment{revisionP}{\par\begingroup\ifrevisionmarks\color{RevisionOrange}\fi}{\par\endgroup}
\definecolor{RevisionRed}{HTML}{C00000}
\newcommand{\revR}[1]{{\ifrevisionmarks\color{RevisionRed}\fi #1}}
\newenvironment{revisionR}{\par\begingroup\ifrevisionmarks\color{RevisionRed}\fi}{\par\endgroup}
\AtBeginDocument{}
\hypersetup{hypertexnames=false}
\defaultbibliographystyle{ACM-Reference-Format}
\defaultbibliography{AfE-v7}

\makeatletter
\AtBeginEnvironment{tcolorbox}{\let\@vspace\@vspace@orig
  \let\@vspacer\@vspacer@orig}
\patchcmd{\ACM@mk@linecount}{\color{red}}{\color{black}}{}{}
\patchcmd{\ACM@mk@linecount}{\color{red}}{\color{black}}{}{}
\newcommand{\UseMainReferences}{%
  \gdef\@extra@binfo{}%
  \gdef\@extra@b@citeb{}%
}
\newcommand{\UseAppendixReferences}{%
  \gdef\@extra@binfo{-app}%
  \gdef\@extra@b@citeb{-app}%
}
\makeatother

\begin{document}

\title{Agents for Experiments, Experiments for Agents:
       A Design Grammar
       for AI-Enabled Experimental Science}

\author{Yingjie Zhang}
\email{yingjiezhang@gsm.pku.edu.cn}
\affiliation{%
  \institution{Guanghua School of Management, Peking University}
  \city{Beijing}\country{China}}

\author{Chun Feng}
\affiliation{%
  \institution{National University of Singapore}
  \city{Singapore}\country{Singapore}}
  \email{chunfeng@u.nus.edu}

\author{Weizhang Zhu}
\affiliation{%
  \institution{Cheung Kong Graduate School of Business}
  \city{Beijing}\country{China}}
  \email{allen2023cu@link.cuhk.edu.hk}

\author{Tianshu Sun}
\affiliation{%
  \institution{Cheung Kong Graduate School of Business}
  \city{Beijing}\country{China}}
\email{tianshusun@ckgsb.edu.cn}

\renewcommand{\shortauthors}{Zhang, Feng, and Sun}

\ccsdesc[500]{Information systems~Information systems applications}
\ccsdesc[500]{Applied computing~Decision analysis}
\ccsdesc[300]{Computing methodologies~Artificial intelligence}

\keywords{AI-for-science; agentic AI; experimental design; condition graphs;
  human-AI workflows; research governance}

\begin{abstract}
\revR{Human-AI experiments vary who receives information, who can act, and
who holds decision authority. Precise descriptions of individual studies
do not by themselves provide a common language for comparing these
arrangements. This conceptual commentary proposes \textsc{SEED} (Structural
Encoding for Experimental Discovery), a grammar that represents each
experimental condition as a typed actor-flow graph. Its purpose is to organize
publicly documented studies into a shared library of consistently represented
conditions. Human researchers and AI systems can use this common reference
to understand prior arrangements and propose extensions for new research
questions. The grammar distinguishes structural changes from attribute
variation; the scientific value of a proposed change requires substantive
argument and evidence. Computational illustrations demonstrate how SEED 
can support design comparison and AI-assisted development of candidate experimental conditions.
We discuss implications for future replication and provenance work, including under
restricted data access, and risks of selective search. We call for community
agreement on the grammar and coordinated construction and maintenance of the
library.}
\end{abstract}

\maketitle

\UseMainReferences
\begin{bibunit}
\section{Introduction}
\label{sec:intro}

\begin{revisionR}
AI systems increasingly advise people, coordinate work, and act alongside
other AI systems \citep{susarla2023janus,swanson2025virtual,ludwig2024machine}.
Their effects depend on task, timing, disclosure, delegation, and user response
\citep{Revilla2023HumanArtificialIntelligen,Wang2025AIAssistantLivestream}.
Experiments provide a foundation for understanding agents and their effects
on people and organizations. By varying who receives information, who can
act, and who retains authority, researchers can test how these arrangements
shape judgment and coordination. This is \textit{experiments for agents}:
using experiments to understand agents and their interactions with humans
and other agents.

Building on these experiments requires connecting findings to the
arrangements each study tests. Explicit design descriptions help researchers
identify what a study changed, compare arrangements, and reconstruct
conditions for further tests, including with different partners or populations.
They also expose commitments that others can scrutinize when proprietary
data or nondisclosure agreements limit access to the underlying evidence,
while leaving the authenticity of inaccessible data unresolved.

AI can assist this effort to build on prior experiments.
In \textit{agents for experiments},
an AI assistant helps researchers encode prior studies, compare arrangements,
and propose candidate conditions for researcher assessment. A shared library
of prior conditions would give researchers and AI systems the same reference
for understanding existing arrangements and exploring extensions. Researchers
could then explain what a proposed experiment retains or changes and why
that change matters for their question.

Both uses require a common basis for describing experimental arrangements.
Precise methods prose describes individual experiments, and factorial tables
organize treatment combinations. Neither convention by itself supplies a
common vocabulary for comparing information routes and decision rights
within conditions. Similar arrangements can appear under different labels;
identically named treatments can involve different actors or authority.
These differences matter both when interpreting experiments about agents
and when using agents to propose further experiments.

This conceptual commentary proposes \textbf{\textsc{SEED}} (Structural
Encoding for Experimental Discovery), a common grammar for organizing publicly
documented studies, regardless of publication status, into a shared library
of experimental conditions. SEED represents each planned or reported condition
as a typed actor-flow graph that identifies actors, their information and
authority relations, and attributes of the actors and flows. An accompanying
design record contains assignment, outcomes, analysis, and other study details.
Library contributors would establish shared reference entries that subsequent
authors reuse when comparing and extending designs. Researchers can use these
records directly or with an AI assistant. Treatment-effect estimates and
evidence of execution remain in the accompanying study documentation.

We develop the grammar through a published investment-advice experiment,
distinguishing changes to actors or flows (\textit{structural mutation}) from
changes to attributes within a fixed structure (\textit{attribute modulation}).
We position these conventions alongside existing representations and reporting
practices, then illustrate their use in design comparison and AI-assisted editing.
We discuss scrutiny, future replication and provenance work, and risks of
selective search and excessive emphasis on novelty, and call for community
agreement on the grammar and collective library construction and maintenance.
\end{revisionR}

\section{\textsc{SEED} as a Conceptual Framework}
\label{sec:framework}

\begin{revisionR}
SEED represents the arrangement tested in one experimental condition.
A graph describes one condition;
an experiment compares conditions. The grammar supplies common rules for
organizing existing conditions into a library and expressing proposed new
conditions in the same language. The accompanying \textit{design record}
links the graphs to the question, task, participants, assignment, outcomes,
analysis, and implementation details. Appendix~\ref{app:condition-graphs} gives formal
definitions; the discussion below develops the concepts needed to use them.
\end{revisionR}

\subsection{Relational Condition Graphs}
\label{sec:expoverview}
\label{sec:advice-example}

A \textit{condition graph} represents one experimental condition.
In \textsc{SEED}, it consists of actors, directed flows between them, and
attributes attached to the actors and flows (see
Appendix~\ref{app:condition-graphs}). We write
\[
G=(V,E),
\]
\revR{where $V$ is the set of actors and $E$ is the set of directed flows.
Different flow types between the same actors are recorded separately.}
For each actor $v\in V$, $\tau_V(v)$ records the actor type and
$\alpha_V(v)$ records actor attributes.
The basic actor types are human actors ($\Delta$) and
\revM{AI actors ($\bigcirc$), including predictive systems and agentic systems.}
Within each actor type, SEED distinguishes roles that differ in function,
information access, or decision rights.
Actor attributes can capture task-relevant capabilities, such as expertise or
model capacity; functional access, such as available external tools; prior
knowledge; task-specific context; and access to dynamic state information.

For each flow $e\in E$, $\tau_E(e)$ records whether the relation is a content
flow ($\rightarrow$), a control flow ($\Rightarrow$), or an iterative
interaction ($\leftrightarrow^n$), and $\alpha_E(e)$ records attributes of
that relation.
\rev{A content flow conveys information, such as advice; a control flow delegates
or returns authority to act; an iterative interaction repeats exchanges.}
\rev{Flow attributes specify the rules attached to an interaction.}
Governance moderators are design parameters that shape the interaction, such
as protocols ($\mathcal{P}$), incentives ($X$), and information design
($\mathcal{I}$).
\rev{Protocols specify how interaction proceeds, incentives specify its rewards
or costs, and information design specifies what actors see.}
Examples include disclosure, masking, confidence thresholds, time limits,
audit requirements, and presentation rules.
These categories are illustrative rather than exhaustive.
Their purpose is to make mechanisms and safeguards inspectable features of a
condition.

Together, these definitions provide a common basis for representing conditions
from publicly documented studies. Researchers can use the same actor, flow,
and attribute conventions to build a shared literature library, linking each
condition graph to its study record and source. Authors would select and cite
agreed entries as the references for comparison and variation, then specify
their proposed changes. New entries and corrections would enter the shared
review and maintenance process described in Section~\ref{sec:governance}.

\vspace{0.3cm}

\begin{revisionM}
\noindent\textbf{A worked example.}
\citet{Yang2025MyAdvisor} test whether human involvement in producing
investment advice changes bank customers' advice-taking. After initial
unaided decisions, customers were randomly assigned to AI-only advice
(control, $G_0$), Human--AI advice (main treatment, $G_1$), or
\textit{Human-only} advice (additional control, $G_2$). The last condition
used the same collaborative advice as $G_1$ without disclosing AI involvement.
The outcome was whether the final investment decision matched the
recommendation.\footnote{\revM{Our reconstruction uses the published article,
Sections 4--5 and endnote 6, and Online Appendix Tables A2--A3,
\url{https://doi.org/10.1287/mnsc.2022.03918}. The graphs are our SEED encoding
of the field study; the separate online experiment in Section 6 is outside
this example.}}
Figure~\ref{fig:advice-conditions} shows the actors, content flows, and selected
attributes. B aggregates bankers with the same advice-production role;
C remains separate because customers retain the investment decision.
Receiving advice does not transfer that authority. Advice was prepared and
selected before delivery; Appendix~\ref{app:governance-encoding} records these
procedures and the remaining study details.
\end{revisionM}

\begin{figure}[htbp]
\centering
\begin{revisionM}
\begin{tikzpicture}[
  ai/.style={draw,circle,minimum size=7mm,inner sep=0pt,font=\small},
  human/.style={draw,regular polygon,regular polygon sides=3,
    minimum size=10mm,inner sep=.3mm,font=\small},
  flow/.style={-{Stealth[length=2.2mm]}},
  role/.style={font=\scriptsize},
  info/.style={font=\scriptsize,align=center,text width=2.8cm},
  every node/.style={align=center}]

\node[font=\scriptsize] at (6.6,1.55)
  {$\bigcirc$ AI actor \qquad $\Delta$ human actor \qquad
   $\rightarrow$ content flow \qquad $\mathcal{I}$ information design};

\node[font=\small\bfseries] at (2.2,.9) {AI-only: $G_0$ (control)};
\node[ai] (a0) at (.6,0) {A};
\node[human] (c0) at (3.8,0) {C};
\node[role,below=2pt] at (a0.south) {AI};
\node[role,below=2pt] at (c0.south) {Customer};
\draw[flow] (a0) -- node[above=5pt,font=\scriptsize] {advice} (c0);
\draw[gray] (2.2,-.08) -- (2.2,-.65);
\node[info] at (2.2,-1.05)
  {$\mathcal{I}$: AI source\\and AI accuracy disclosed};

\draw[gray!50] (4.8,1.05) -- (4.8,-1.5);
\node[font=\small\bfseries] at (9.2,.9) {Human--AI: $G_1$ (treatment)};
\node[ai] (a1) at (6,0) {A};
\node[human,line width=1.1pt] (b1) at (9.2,0) {B};
\node[human] (c1) at (12.4,0) {C};
\node[role,below=2pt] at (a1.south) {AI};
\node[role,below=2pt] at (b1.south) {Banker role};
\node[role,below=2pt] at (c1.south) {Customer};
\draw[flow,line width=1.1pt] (a1) --
  node[above=5pt,font=\scriptsize] {risk assessment} (b1);
\draw[flow,line width=1.1pt] (b1) --
  node[above=5pt,font=\scriptsize] {selected advice} (c1);
\draw[gray] (10.8,-.08) -- (10.8,-.65);
\node[info] at (10.8,-1.05)
  {$\mathcal{I}$: banker + AI source\\and AI accuracy disclosed};

\node[font=\small\bfseries] at (6.5,-2.05)
  {\textit{Human-only}: $G_2$ (additional control)};
\node[ai] (a2) at (3.3,-3.05) {A};
\node[human] (b2) at (6.5,-3.05) {B};
\node[human] (c2) at (9.7,-3.05) {C};
\node[role,below=2pt] at (a2.south) {AI};
\node[role,below=2pt] at (b2.south) {Banker role};
\node[role,below=2pt] at (c2.south) {Customer};
\draw[flow] (a2) -- node[above=5pt,font=\scriptsize] {risk assessment} (b2);
\draw[flow] (b2) -- node[above=5pt,font=\scriptsize] {selected advice} (c2);
\draw[gray] (8.1,-3.13) -- (8.1,-3.7);
\node[info,draw,dashed,inner sep=4pt] at (8.1,-4.15)
  {$\mathcal{I}$: banker source;\\AI involvement and\\AI accuracy not disclosed};

\draw[gray!60] (0,-4.95) -- (13.2,-4.95);
\node[anchor=north west,font=\scriptsize] at (0,-5.1)
  {\textbf{Node attributes $\alpha_V$} (apply wherever the role appears)};
\node[anchor=north west,align=left,font=\scriptsize,text width=3.8cm]
  at (0,-5.6) {A: task = loan-risk assessment};
\node[anchor=north west,align=left,font=\scriptsize,text width=4.3cm]
  at (4.1,-5.6) {B: AI access = risk assessments;\\
  no AI investment recommendations};
\node[anchor=north west,align=left,font=\scriptsize,text width=4cm]
  at (9,-5.6) {C: authority = final\\investment decision};
\node[anchor=north west,align=left,font=\scriptsize,text width=13.2cm]
  at (0,-6.55) {\textbf{Flow attributes $\alpha_E$:}
  $\mathcal{I}$ is shown beside each advice-delivery edge.
  For B $\rightarrow$ C, protocol $\mathcal{P}$ selects the most frequent
  advice for each loan before delivery.};
\end{tikzpicture}
\end{revisionM}
\caption{\revM{Our role-level encoding of the field experiment.
Circles denote AI and triangles humans. Annotations show node attributes
and flow attributes. Thick edges mark the mediated route; the dashed box
marks source-information differences in $G_2$. }}
\label{fig:advice-conditions}
\Description{Three condition graphs use circles for AI and triangles for
human bankers and customers. AI-only connects A to C; Human-AI and Human-only
connect A to B to C. Node attributes describe A's task, B's AI information
access, and C's decision authority. Edge attributes describe the source
information disclosed to customers and the advice-selection protocol.
Human-only retains AI in production but omits AI involvement and accuracy
from the source description.}
\end{figure}

\subsection{Varying Conditions}
\label{sec:architect}

To propose new designs, researchers and AI systems first encode a baseline
condition or select one from the library, then vary this reference.
The research question motivates which features of the
reference graph $G_0$ to change. Applying an operation $\mathcal{O}$ produces
an alternative condition: $G_1=\mathcal{O}(G_0)$.

\textsc{SEED} distinguishes two broad classes of such operations.
\textit{Structural mutation} changes the actor-flow structure itself: it adds
or removes actors, changes connectivity, alters flow type, introduces iterative
exchange, or reverses authority.
For example, a design may insert a human reviewer between an AI system and a
downstream decision maker, or reverse a control flow from
human-delegates-to-AI ($\Delta \Rightarrow \bigcirc$) to
AI-escalates-to-human ($\bigcirc \Rightarrow \Delta$).
These changes are mechanism-relevant because they alter who observes, acts,
or controls the workflow.

\textit{Attribute modulation} keeps the actor-flow structure stable while
changing actor or flow attributes.
Examples include varying AI capability, toggling identity disclosure, masking
the source of advice, changing incentives, \revM{varying an existing audit requirement}, or
calibrating a confidence threshold in a protocol moderator ($\mathcal{P}$).
If a reference condition already permits both direct AI advice and advice
through a human reviewer, changing the threshold for requesting review is
attribute modulation. Adding a previously unavailable reviewer and review
route is structural mutation.
This distinction matters because two conditions may share the same actor-flow
structure while differing substantively in the rules that govern interaction.
Conversely, a small-looking change in authority or feedback can \revN{alter a
hypothesized mechanism}.

\revP{Each actor, flow, and attribute needs a defensible empirical meaning,
and each graph change should correspond to a feasible contrast in roles,
information, authority, feedback, or safeguards. These requirements keep
comparison and generation tied to the research question and setting.
The grammar makes proposed changes explicit; their novelty and scientific
value require substantive argument.}
Conditions map to graphs and manipulations to graph changes; assignment,
sample, outcomes, and analysis remain in the accompanying record.

\vspace{0.3cm}

\begin{revisionM}
\noindent\textbf{An illustrative example.}
In Figure~\ref{fig:advice-conditions}, take the AI-only control $G_0$ as a
library reference: A sends advice directly to C. Reconstruct $G_1$ by inserting
human banker B and replacing $A\rightarrow C$ with $A\rightarrow B\rightarrow C$.
This is \textit{structural mutation}: the actor set and connections change.
The reported contrast also changes advice content and source information,
so it bundles structural and attribute changes.
Next, obtain $G_2$ from $G_1$ by keeping the actors, route, and advice content
fixed while changing the information-design attribute $\mathcal{I}$ on
$B\rightarrow C$: the source description omits AI involvement and accuracy.
This is \textit{attribute modulation}: an existing source-information
attribute changes value while the actor-flow structure stays fixed.\footnote{These operations reconstruct published conditions that may already be
library entries.}
Researchers and AI systems can use these encodings to propose further
conditions, whose novelty requires comparison with the library
(Section~\ref{sec:os}).
\end{revisionM}

\begin{revisionR}
\subsection{Relationship to Existing Representations}
\label{sec:positioning}

SEED complements familiar descriptions of experiments. Methods prose reports
procedures and a factorial table organizes treatment combinations; condition
graphs make information flows and decision authority explicit within each
condition. Common role and flow categories support comparison across studies.
For example, a search for mediated AI advice would include the worked example's
\textit{Human-only} condition because AI remains in its production route.

Business Process Model and Notation (BPMN), Petri nets, and W3C PROV already
provide languages for processes, workflows, and provenance
\citep{OMGBPMN2014,VanDerAalst1998Workflow,W3CPROV2013}.
SEED's contribution is a set of conventions for experimental comparison:
one graph per
condition, separate information and authority relations, function-based role
matching, and explicit changes between conditions. These conventions could
extend an existing language, with comparisons using compatible scope and
role granularity.

Structured reporting and registration address what to document and when to
commit. TIDieR calls for reporting intervention procedures, providers, and
planned and actual delivery \citep{Hoffmann2014TIDieR}. Preregistration records
prospective commitments; registered reports add review before outcomes are
known and in-principle publication acceptance
\citep{COSPreregistration,COSRegisteredReports}. SEED supplies a common
language for conditions within these records. The same language can describe
reported conditions retrospectively, as in Appendix~\ref{app:governance-encoding}.
\end{revisionR}

\section{Design Functions of \textsc{SEED}}
\label{sec:os}

\revR{SEED connects three uses: describing existing conditions in a shared
library, comparing their structures, and proposing new designs. The library
provides the common reference; researchers and AI systems can develop
different extensions and argue for their scientific contribution.}

\subsection{Function 1: Describing Experimental Conditions}
\label{sec:descriptive}

First, \textsc{SEED} gives researchers a compact language for describing the
experimental condition, not only the topical domain.
A condition can be encoded by naming its actors, typing the flows among them,
and attaching the attributes that govern visibility, authority, timing, and
feedback.
Many human-AI studies then appear as variants of a few recurring actor-flow
families: AI advises a human ($\bigcirc \rightarrow \Delta$), a human
delegates to AI ($\Delta \Rightarrow \bigcirc$), AI escalates back to a human
($\bigcirc \Rightarrow \Delta$), or both parties interact iteratively
($\bigcirc \leftrightarrow^n \Delta$).
Table~\ref{tab:Literature_mapping} provides representative examples. 
\revN{Separating the actor-flow structure from the domain narrative allows
researchers to identify similar arrangements for substantive comparison
across settings.}

\begin{table}[t]
\centering
\caption{Representative Human-AI Designs as \textsc{SEED} Actor-Flow Families.}
\label{tab:Literature_mapping}
\renewcommand{\arraystretch}{1.35}
\setlength{\tabcolsep}{5pt}
\begin{tabular}{p{0.18\textwidth} p{0.22\textwidth} p{0.5\textwidth}}
\hline
\textbf{Structure} & \textbf{Design Family} &
\textbf{Representative Examples}\\ \hline

$\bigcirc \rightarrow \Delta$ &
AI advice or assistance &
\revM{Selected AI-to-human information flows in advice or assistance}
\citep{You2022AlgorithmicVersusHuman, Xu2024VoiceChatbotAnthropomorphism,
Brynjolfsson2025GenerativeAiAt, Goh2025Gpt4Assistance} \\

$\Delta \Rightarrow \bigcirc$ &
Delegation to AI &
Human actors delegate task execution or decision authority to AI systems
\citep{Fuegener2022CognitiveChallengesIn} \\

$\bigcirc \Rightarrow \Delta$ &
Escalation to human &
AI systems return authority or unresolved cases to human actors
\citep{Stelmaszak2025WhenAlgorithmsDelegate} \\

$\bigcirc \leftrightarrow^n \Delta$ &
Iterative collaboration &
Human and AI actors exchange information across multiple interaction rounds
\citep{Chen2025LargeLanguageModel, Lin2024DecisionOrientedDialogue} \\

$\bigcirc \rightarrow \Delta_I \rightarrow \Delta_D$ &
Mediated AI advice &
\revM{Bankers ($\Delta_I$) use AI risk assessments to form advice;
the study selects advice for downstream customers ($\Delta_D$)}
\citep{Yang2025MyAdvisor} \\
\hline
\end{tabular}
\Description{The table maps five representative human-AI research designs to
SEED actor-flow families, including AI advice, delegation, escalation,
iterative collaboration, and mediated AI advice.}
\end{table}

\revM{In the voice-chatbot field experiment by
\citet{Xu2024VoiceChatbotAnthropomorphism}, identity disclosure and
anthropomorphism vary attributes of a human--chatbot interaction. Drivers
respond to the chatbot; the single AI-to-human arrow in
Table~\ref{tab:Literature_mapping} represents only the outbound communication.}
By contrast, the financial-advisory field experiment by \citet{Yang2025MyAdvisor}
\revM{provides the direct and mediated production routes encoded in
Section~\ref{sec:advice-example}, together with source-information variation.
The route comparison does not imply that all other treatment features are fixed.}
Both designs are substantively important, but they occupy different positions
in the design space.
\rev{Library queries group studies by their experimental arrangements;
researchers can use these groups alongside syntheses of findings.}

\subsection{\revR{Function 2: Comparing Design Structures}}
\label{sec:novelty}

\revR{SEED compares a proposed condition with encoded reference conditions.
The comparison separates changes in the interaction arrangement from changes
in its parameters or context. A combined distance can be written as:}
\begin{equation}
    D(G_{new}, G_{ref}) = w_s \cdot \delta_{struct}(G_{new}, G_{ref})
                        + w_p \cdot \delta_{param}(G_{new}, G_{ref}).
\end{equation}
Here, $\delta_{struct}$ captures differences in the actor-flow skeleton or in
\rev{the presence of attribute categories, called slots,} such as adding an actor, reversing
authority, introducing a feedback loop, or adding an incentive moderator.
$\delta_{param}$ captures differences within matched slots or contextual
descriptors, such as the specific incentive rule, disclosure wording,
confidence threshold, domain, stakes, urgency, or information asymmetry
\citep{Puranam2021HumanAIDecision,
Rastogi2022DecidingFastSlow, Vossing2022DesigningTransparency, Kawakami2022ChildWelfare,
XuZhang2022ContextTheorizing}.
\revR{A slot is a substantive design-attribute category, not a documentation
field or an outcome measurement. Disclosed and undisclosed sources occupy
the same source-information category. An existing audit requirement can
vary parametrically, while adding an oversight role changes the structure.
The library specification fixes these categories before comparison and
distinguishes unreported information from confirmed absence. For example,
changing a reward from zero within an existing incentive category is
parametric; adding a previously absent incentive category is structural.
Apply the same convention to reference and proposed conditions.}
\rev{The researcher-chosen weights} $w_s$ and $w_p$ encode design priorities: a field seeking
new mechanisms may weight structure more heavily, while a field seeking
generalizability may weight contextual variation more heavily.
Appendix~\ref{app:structural-novelty} connects the structural term to graph
edit distance\rev{, the minimum cost of edits that transform one graph into
another,} and explains how raw edit costs can be scaled to $[0,1]$.

\begin{revisionR}
Using the library's scope, role definitions, and categories, match roles by
function and decision rights, consulting study records for substantive
equivalence. Distances help locate related entries and identify what a
proposed condition changes. The scientific value of structural or parametric
changes requires substantive argument.
\end{revisionR}

\subsection{Function 3: Generating Candidate Designs}
\label{sec:generative}

Researchers and AI systems can propose new conditions from the same library.
The proposer must explain how a research situation maps to a candidate graph,
what it changes relative to existing entries, and why those changes warrant
study. Different proposals and assessments of novelty or research value can
arise from this common base.

AI-assisted generation is the core motivation for \textit{agents for
experiments}.
\rev{This design assistant works outside the experimental arrangement; it is
distinct from the AI advisor whose effects the example tests.}
Given a research question, fixed design elements, domain constraints, and
governance boundaries, an AI system can retrieve related condition graphs,
apply permitted structural mutations or attribute modulations, and return
candidate designs that make the underlying actor-flow change explicit (see
Appendix~\ref{app:prompt-scaffold}).
Researchers assess the candidates' scientific novelty, operational
feasibility, and governance risks, then select, revise, or reject them.
They retain responsibility for any design that moves forward.

\revP{Candidate generation should use available data, permissions, recruitment
capacity, and implementation resources as constraints. Researchers should
assess whether they can enact the manipulation, observe required outcomes,
and answer the question through the proposed comparison. The graph and record
expose these requirements; methodological assessment determines whether
implementation is warranted.}


\section{\revR{Illustrating Comparison and Design Editing}}
\label{sec:demo}

\begin{revisionR}
Two illustrations show how to compare conditions with prior designs in a
library and how to represent and check design changes across different formats.

\subsection{Locating a Change Relative to Prior Designs}
\label{sec:demo1}

A researcher extending an existing condition needs to identify what changes
and whether those features already appear in prior designs. We illustrate
this comparison with a library of eight reference conditions from four
studies and twelve constructed variants
\citep{You2022AlgorithmicVersusHuman,Yang2025MyAdvisor,
Revilla2023HumanArtificialIntelligen,Goh2025Gpt4Assistance}.

We first compare each variant with its starting condition, its parent.
Changing identifiers or record order leaves the distances at zero. Changing
existing attributes or task context enters the parameter component; changing
roles or relations enters the structural component. The decomposition makes
the nature of the proposed change explicit.

We then compare the candidate with the whole library to locate related prior
conditions. For example, removing the banker from an AI-to-banker-to-customer
arrangement creates direct AI-to-customer advice. This changes the parent
structure, but the resulting arrangement already appears in the library.
Parent distance is positive while nearest-library distance is zero on the
encoded, comparable features. The retrieved record thus gives the researcher
a prior design to examine before treating the proposed change as new.

Interpreting these matches requires inspecting the comparison rules and
available attributes. Changing weights or scaling changes the nearest
references for two of twelve queries; changing edit costs can also change
the minimum-cost role correspondences, including ties. Missing attributes
limit what a zero distance means. Appendix~\ref{app:novelty-coding} traces
these choices and results. The constructed examples illustrate comparison
and retrieval. Agreement with their assigned change categories checks that
the calculation implements the specified edits.

\subsection{Editing Conditions Across Representations}
\label{sec:demo2}

The second illustration examines AI-assisted condition editing and study-plan
specification using detailed prose, a BPMN-inspired table, and SEED records.
The three formats express the same task requirements: prose connects the
facts in sentences, the table organizes actions and handoffs, and SEED
separates roles, relations, and changes. All can specify authority and
conditional interaction; the process table is not a full BPMN implementation.

A model proposes edits and a study plan. A shared interpreter converts the
edits to a common record, retaining unchanged relations, before checks and
feedback. After this support, all three formats pass the sixteen held-out
instances and twelve exploratory stress tasks. The held-out instances vary
roles or parameters within eight developed mechanism families.
Appendix~\ref{app:anchors-operators} reports the procedure and initial and
revised checks. These checks establish whether the requested changes are
implemented while preserving the specified relations and final authority.

A study plan must also specify how to compare conditions, measure outcomes,
analyze data, and carry out the study. Supplementary model ratings assess
these aspects of the forty-eight final held-out plans, sixteen per format.\footnote{The
generation model also rates the plans, using a project-defined rubric.
Appendix~\ref{app:rubric} details the scoring procedure and its limitations.}
SEED has higher mean ratings on all four dimensions: treatment contrast,
measurement, analysis, and implementability. Its mean total is 12.875 out of
16, compared with 8.563 for prose and 10.688 for the process table
(Appendix~\ref{app:rubric}--\ref{app:quality-results}).

The ratings indicate an advantage in how fully SEED plans specify the study
within this batch. This provides preliminary support for using explicit
roles, relations, and changes to guide AI-assisted design development.
\end{revisionR}

\section{\revR{Considerations for Using SEED}}
\label{sec:governance}

\begin{revisionR}
A common design language can aid comparison while also changing what
researchers search for and report. Three tensions matter for its use.

\subsection{Novelty vs.\ Replication}

Structural comparison may draw attention to new arrangements at the expense
of needed replications. Yet replication and contextual re-testing are central
to cumulative science \citep{OpenScienceCollaboration2015,NationalAcademies2019}.
Changes in models, interfaces, and user expectations make them especially
relevant to experiments involving agents.

SEED's common language could inform future replication tools for documenting
retained and changed roles, routes, and authority and locating designs in a
shared library when original data are inaccessible. Classifying direct replication, contextual re-testing,
or tests of different mechanisms requires substantive judgment. Tool
development and evaluation remain future work.

Model replacement can preserve an arrangement while changing behavior;
retiring a model may prevent strict reproduction. Recording versions makes
this constraint visible. Replication incentives also depend on research and
editorial priorities. Researchers and editors should assess such work by the
uncertainty it resolves: zero structural distance, or even zero overall
novelty, can accompany a valuable test of robustness.

\subsection{Speed vs.\ Validity}

Agents can propose conditions faster than researchers can assess feasibility
and validity. Human responses shape outcomes alongside model accuracy
\citep{Amershi2019GuidelinesHumanAI,Bansal2019BeyondAccuracyMentalModels}.
Researchers must assess whether the manipulation, measures, data access,
and sample support the intended inference. Accessible data and simulations
can ease implementation, but simulation findings depend on modeled behavior
and assumptions.

SEED organizes roles, routes, and decision rights in common categories, with
activation and ordering rules in the accompanying record. These commitments
give reviewers specific points to check under restricted data access.
Precise prose can state them too; SEED proposes consistent organization for
inspection and comparison.
If advice delivery requires human approval, a disclosed summary could identify
deliveries without recorded prior approval. Such cases warrant questions
about protocol deviations or incomplete reporting. Without supporting
information, the encoding identifies what to request. Authenticating data
and verifying execution require evidence beyond consistent self-reports.

The design record could inform future provenance infrastructure through links
to dataset hashes at delivery, partner attestation, or confidential auditor
access. Verification procedures require separate technical and institutional
development and evaluation; their evidential value depends on supporting
materials and checks.

Faster generation also lowers the cost of selective search. Researchers can
test many variants against pilot outcomes and report only the successful
condition; its final graph conceals that history. Distinguish exploration
before data contact from pilot-informed development, and document subsequent
changes and rejected alternatives. Timestamped preregistration and amendments
can expose prospective commitments \citep{COSPreregistration}; journals or
institutions could request these records. Self-reporting can nevertheless
omit searches and prior data contact. Generation can also be optimized to
satisfy a screen's stated criteria. Complete, consistent fields establish
neither validity nor a complete search history.

\subsection{Standardization vs.\ Diversity of Inquiry}

SEED is most useful when a claim depends on actors, information, authority,
or feedback. It adds less to already transparent contrasts and is less suited
to open-ended or interpretive inquiry whose roles emerge during the study.
Infrastructure that favors easy-to-encode designs could narrow attention to
tractable, data-rich questions \citep{bhargava2025exploring}.
Researchers, reviewers, and funders should ask what inquiry is left outside
the grammar.

A shared library requires community agreement on conventions and records.
Contributors would jointly check entries against public sources, document
reasons for role grouping, flow types, and attribute categories, and discuss
differing interpretations. Source checks and clarification of the conventions
would guide agreement; unresolved features remain marked. Authors extending
a documented study would cite and reuse its agreed library entry, specify
their proposed condition, and explain the changes. New entries and corrections
would follow the same shared review process. Each version should retain its
encoding rationale so users can identify the reference behind a comparison
or new design. The grammar can develop through community input, including
from users of existing workflow standards. Studies of researcher use can
assess the library's practical benefits.

Across these tensions, preserve the reference condition, proposed change,
assessment, rejected alternatives, and selection rationale when SEED or an
AI assistant shapes a design. Appendix~\ref{app:governance-encoding} illustrates
the study documentation through a retrospective case record. Documented choices
support scrutiny while
researchers retain responsibility for the work \citep{susarla2023janus}.

Different readers can use this common language for different judgments.
Researchers can relate an edit to a mechanism; reviewers can compare the
claim with the actual treatment differences. Institutional review boards
(IRBs) can inspect information access, disclosure, and decision authority
when assessing participant risks. Editors can identify supporting records
or confidential verification to request for restricted-data studies.
For replicators, the common categories could guide assessment of what to
retain as models and settings change. The common record connects these
assessments to the same design specification.
\end{revisionR}

\section{Conclusion}
\label{sec:conclusion}
\begin{revisionR}
SEED proposes a common grammar for organizing publicly documented studies
into a shared library of experimental conditions that researchers and AI
systems can understand, compare, and extend. The worked case and illustrations
explain this use; a new graph's scientific value depends on the question,
mechanism, and evidence behind it.

This connects experiments for understanding agents with agents that assist
experimental design. The common language also offers a basis for future
replication and provenance work under restricted data access. Developing
those tools and assessing practical benefits require further work.

If the community adopts these conventions, the next priority is to build and
maintain the library. We call on researchers to contribute condition records
from publicly available studies, regardless of publication status, and to
keep the shared specification, source links, and records current. This
collective effort would give human and AI inquiry a common, growing base
from which to develop new experiments.
\end{revisionR}

\renewcommand{\refname}{References}
\putbib
\end{bibunit}

\clearpage
\appendix
\setcounter{page}{1}
\makeatletter
\@addtoreset{table}{section}
\@addtoreset{figure}{section}
\makeatother
\renewcommand{\thetable}{\thesection\arabic{table}}
\renewcommand{\thefigure}{\thesection\arabic{figure}}
\UseAppendixReferences
\begin{bibunit}
\section{Technical Appendix: \textsc{SEED} Representation and Design Functions}
\label{app:formal}

We provide the formal details that support the main-text framework.
The appendix is intentionally more technical than the commentary body: it
defines the condition-graph representation, specifies the novelty score, and
provides \rev{reusable prompt instructions} for design generation.

\subsection{Attributed Relational Condition Graphs}
\label{app:condition-graphs}

The representation defined here supplies the common grammar for condition
records in the shared literature library. A \textsc{SEED} condition is
represented as a typed, attributed, directed relational graph
\[
G = (V,E),
\]
\revR{Here $V$ is the set of actors and $E$ is the set of directed flows.
Each flow has a source, recipient, and type; distinct typed relations may
connect the same ordered pair of actors.}
We use ``flow'' as the prose term for a graph edge because the relations of
interest are information, authority, and interaction flows; formally, a flow
is an element $e\in E$.
The graph is relational, not computational: it represents who participates in
a condition, what information or authority moves among them, and how the
rules of interaction are attached to those relations.
Actors and flows carry different kinds of information, so we distinguish
actor-level and flow-level type and attribute functions.
For each actor $v\in V$, $\tau_V(v)$ records the actor type and
$\alpha_V(v)$ records actor attributes.
For each flow $e\in E$, $\tau_E(e)$ records the flow type and $\alpha_E(e)$
records flow attributes.

\paragraph{Actors.}
Each actor $v \in V$ represents a role-bearing entity in the condition.
Its type is
\[
\tau_V(v) \in \{\Delta,\bigcirc\},
\]
where $\Delta$ denotes a human actor and $\bigcirc$ denotes \revN{an AI
actor, including predictive and agentic systems}.
Subscripts can be added to indicate roles within a condition, such as
$\Delta_I$ for a human intermediary, $\Delta_R$ for a human reviewer,
$\Delta_D$ for a downstream decision maker, or $\bigcirc_A$ for an AI
assistant.
The human/AI distinction is intentionally minimal.
It marks whether a role is occupied by a human actor, whose behavior may
reflect expertise, incentives, cognitive limits, and accountability
obligations, or by \revN{an AI actor}, whose behavior is shaped by model
capabilities, instructions, tools, memory, and access constraints.

\rev{Actor attributes record task-relevant capabilities and access.
\textit{Cognitive capability} concerns reasoning capacity relevant to the
task, such as a human reviewer's clinical expertise, a participant's
statistical literacy, or an AI actor's model capability and reasoning depth.
\textit{Functional capability} concerns the availability and permitted use
of external tools. For humans, these may include databases, dashboards,
communication channels, calculators, or decision-support systems.
For AI actors, they may include retrieval, code execution, web search,
database queries, API calls, or simulation tools that extend what the actor
can observe or do
\citep{Schick2023Toolformer, Patil2024Gorilla, Lewis2020RAG}.}
\rev{\textit{Prior knowledge} is relatively stable knowledge brought to the
task, such as professional training for humans or pretraining and
domain-specific fine-tuning for AI actors. \textit{Task-specific context}
is information supplied in the condition, such as a case vignette, data
table, patient history, customer transcript, or research brief.
\textit{Access to dynamic state information} concerns what an actor can
observe as the task unfolds, such as real-time workload, updated sensor
readings, recent interaction history, or other actors' current workflow
status. The encoding should retain the attributes needed to interpret the
experimental contrast, as with the banker's access to AI risk assessments
in the worked example.}

\paragraph{Flows.}
\revR{Write $e_{ij}$ for a flow from actor $v_i$ to actor $v_j$, with
the subscript indicating its endpoints rather than uniquely identifying
the relation.}
Its type is
\[
\tau_E(e_{ij}) \in \{\rightarrow,\Rightarrow,\leftrightarrow^n\}.
\]
\begin{itemize}
    \item \textit{Content flows ($\rightarrow$)} \rev{transmit information}, such as data, predictions, rationale, or generated text.
    Traversing this flow updates the receiver's information state but does not
    force action.
    \item \textit{Control flows ($\Rightarrow$)} transmit executive authority,
    such as triggers, vetoes, or final decision tokens, that may update the
    receiver's action state.
    \item \textit{Iterative interactions ($\leftrightarrow^n$)} are
    bidirectional exchanges over $n$ rounds. The notation distinguishes a
    one-shot handoff from a repeated or co-evolutionary interaction.
\end{itemize}
\revN{The edge set records routes permitted by the study protocol within the
declared scope, annotated as required, optional, or conditional. A realized
trace records the interactions that occur in a particular case. The
accompanying record specifies triggers, ordering, and stopping rules;
unreported routes remain unresolved rather than being coded as absent.}
Formally, $v_i \leftrightarrow^n v_j$ can be represented as reciprocal
directed flows with an iteration attribute, or as a time-indexed graph
sequence when the interaction changes over time.
When the interaction structure itself changes over time, the condition can be
represented as a graph sequence $\{G_t\}_{t=1}^{n}$.
\revN{If this sequence specifies planned stages, record the transition rules.
A sequence reconstructed from execution instead describes realized behavior.
Neither use of the notation by itself specifies a distribution over traces.}

\paragraph{Governance Moderators.}
Governance moderators are \rev{rules specified in the design} and attached to actors or
flows depending on what they govern.
The following three categories are typical rather than exhaustive.
For compactness, we present the flow-level form, which covers the main
examples in the commentary. For a flow $e$, these can be written as
\[
\alpha_E^{gov}(e)=\{\mathcal{P}_e,X_e,\mathcal{I}_e\}.
\]
\begin{itemize}
    \item \textit{Protocols ($\mathcal{P}_e$)} determine the admissibility of a flow. 
    These are hard constraints governing when a signal is valid. 
    Common implementations include confidence thresholds 
    (e.g., ``Delegate ONLY IF Confidence $>$ 0.8''), time or resource limits 
    (e.g., ``Veto allowed within 500ms''), and format constraints 
    (e.g., ``Output must be structured JSON'')
    \citep{Mozannar2020ConsistentEstimatorsDefer}.
    \item \textit{Incentives ($X_e$)} define payoff-relevant rules that shape
    preferences over possible outcomes
    \citep{koster2022human,thaler2021nudge}.
    \item \textit{Information design ($\mathcal{I}_e$)} governs what is
    revealed and how it is presented, including transparency, masking,
    identity disclosure, explainability, and nudges
    \citep{Zhang2020ConfidenceExplanationTrust,
    Shi2020PersuasiveDialoguesBotIdentity, Chan2024VisibilityAIAgents}.
\end{itemize}

\paragraph{\rev{Interaction Dynamics.}}
\begin{revisionM}
We use ``interaction dynamics'' to refer to endogenous relational states that
can emerge, change, or be measured as actors interact.
They are usually attached to flows because they describe the evolving quality
of a connection between actors.
The following dimensions are illustrative rather than exhaustive.
\begin{itemize}
    \item \textit{Psychological alignment} captures affective
    relation quality, including trust and psychological safety
    \citep{Yin2019AccuracyTrust, Bansal2019BeyondAccuracyMentalModels}.
    \item \textit{Epistemic alignment} captures informational
    consensus or disagreement between actors
    \citep{Zhang2020ConfidenceExplanationTrust,
    Bansal2019BeyondAccuracyMentalModels}.
    \item \textit{Cognitive alignment} captures process
    efficiency, including mental workload or computational latency
    \citep{Amershi2019GuidelinesHumanAI, Kocielnik2019ImperfectAIExpectations}.
\end{itemize}
\end{revisionM}

\subsection{Structural Distance and Novelty}
\label{app:structural-novelty}

Let $\mathcal{L}=\{G_1,\ldots,G_n\}$ denote a library of encoded prior
condition graphs.
For a new condition graph $G_{new}$ and a reference condition graph
$G_{ref}$, define:
\[
D(G_{new}, G_{ref}) =
w_s \cdot \delta_{struct}(G_{new}, G_{ref})
+ w_p \cdot \delta_{param}(G_{new}, G_{ref}).
\]
The components are normalized so that
$\delta_{struct},\delta_{param}\in[0,1]$, with
$w_s,w_p\geq 0$ and $w_s+w_p=1$.
The structural component compares the actor-flow skeleton and the presence of
\rev{attribute categories (slots).}
It asks whether two conditions instantiate the same relational design: the
same actors, flows, flow types, authority relations, feedback loops, and
\revN{design-attribute categories, such as moderators. Measurements of
realized trust, effort, or other states do not themselves count as structural
edits; an associated change to the protocol is classified by the design
features it alters.}
It can be operationalized as a weighted graph edit distance followed by a
transparent scaling rule.  Let the raw edit cost be:
\[
d_{\mathrm{edit}}(G_a,G_b)=
\min_{\pi\in\Pi(G_a,G_b)}\sum_{o\in\pi}c(o),
\]
where $\Pi(G_a,G_b)$ is the set of valid edit paths that transform one
condition graph into another and $c(o)$ is the researcher-specified cost of edit
operation $o$ \citep{Ranjan2022GREED, Jain2024GraphEdX}.
The structural distance is then the scaled edit cost:
\[
\delta_{struct}(G_a,G_b)=T_q\!\left(d_{\mathrm{edit}}(G_a,G_b)\right),
\]
where $T_q$ maps raw edit costs to $[0,1]$ for the study context $q$.  For
example, if costs are not already bounded, one can use
$T_q(x)=\min(1,x/d_{\max})$, where $d_{\max}>0$ is a design-specified upper
bound for the relevant comparison set.
\revN{In the feasibility illustration, Appendix~\ref{app:novelty-coding}
assigns raw costs to edits and uses the fixed display scale of 8, giving
$T_q(x)=\min(1,x/8)$.}
In \textsc{SEED}, valid structural operations include adding or deleting an
actor, adding or deleting a flow, redirecting a flow, reversing authority,
changing a flow type, adding an iterative interaction, or adding or removing
an attribute slot.
The cost function should be theory-driven and auditable: adding a disclosure
\revN{category would receive an explicit cost under the chosen convention,
as would adding a reviewer or changing authority. Disclosed and undisclosed
values within an existing category are parametric alternatives. These costs
make the comparison's modeling judgments explicit.}

The parametric component compares values or specifications within matched
structural slots, along with contextual descriptors selected for the study.
For example, two condition graphs may both include an incentive moderator
$X_e$ on the same flow; the presence of that moderator belongs to the shared
structure, while differences between a speed bonus, an accuracy bonus, or a
tournament payment rule belong to $\delta_{param}$.
The same logic applies to disclosure wording, confidence-threshold levels,
model versions, tool availability, participant expertise, empirical domain,
stakes, urgency, or interaction medium.
Formally, for a study-specific set of relevant parameters $\mathcal{K}_q$, one
can write:
\[
\delta_{param}(G_a,G_b)=
\operatorname{Agg}_{k\in\mathcal{K}_q} d_k(G_a,G_b),
\]
where each $d_k(G_a,G_b)\in[0,1]$ records the degree of difference on parameter
$k$, and $\operatorname{Agg}$ is a transparent aggregation rule, such as a mean
or weighted mean.
The key requirement is that each feature be assigned to one component before
scoring.
Slot presence is structural; variation within an already matched slot is
parametric.
\revM{Apply this convention to substantive categories shared across the
comparison (Section~\ref{sec:novelty}). Documenting a previously
unreported category does not establish that the underlying design changed.
Disclosed and undisclosed values of an existing source-information category
are parametric alternatives, not the addition or deletion of AI from production.}
\revN{Declare category meanings before comparison. If reasonable encodings
disagree about category presence or value, document the alternatives and their
implications rather than attributing the resulting distance to the study alone.
Use compatible scope and role correspondence across the reference set.}
Reporting $\delta_{struct}$ and $\delta_{param}$ separately lets readers
distinguish actor-flow novelty from parameter variation and contextual
transfer.

The novelty score of a proposed design is its distance to the nearest prior
condition graph in the library:
\[
\mathcal{N}(G_{new}\mid \mathcal{L}) =
\min_{G_\ell \in \mathcal{L}} D(G_{new},G_\ell).
\]
Thus, $\mathcal{N}=0$ when a condition graph exactly matches a prior encoded
condition, and larger values indicate greater distance from the nearest prior
condition graph.
A design-generation system can then rank a finite set of candidate condition
graphs generated under a research question $q$ and constraints $c$:
\[
G^*\in\operatorname*{argmax}_{G\in\mathbb{G}_{\mathrm{cand}}(q,c)}
\left[\mathcal{N}(G\mid\mathcal{L})-\lambda\mathcal{C}(G)\right],
\]
where $\mathbb{G}_{\mathrm{cand}}(q,c)$ is the candidate set,
$\mathcal{C}(G)$ is a normalized complexity or feasibility penalty, and
$\lambda\geq0$ is \rev{a researcher-chosen weight on that penalty}.
This is a ranking heuristic over generated candidates, not a claim that the
system can exhaustively optimize over all possible experimental designs.
Operational feasibility or coherence scores used in pilots should be treated
as implementation heuristics, not as core mathematical primitives.

\subsection{Generative Prompt Scaffold}
\label{app:prompt-scaffold}

\begin{figure}[H]
\centering
\fbox{
\begin{minipage}{0.94\textwidth}
\small
\textbf{Prompt Scaffold: \textsc{SEED} Design Agent}

\vspace{0.15cm}
\textbf{Role:} \revP{Use the \textsc{SEED} grammar to propose candidate
experimental conditions and identify questions for researcher assessment.}

\textbf{Inputs:}
\begin{itemize}
    \setlength\itemsep{0em}
    \item Condition-graph library $\mathcal{L}$ of encoded prior studies.
    \item Research question $q$, fixed design elements, and target domain.
    \item Feasibility and governance constraints $c$, such as high stakes,
    time pressure, audit requirements, or IRB-sensitive information.
    \item Permitted operations: structural mutation or attribute modulation.
\end{itemize}

\textbf{Task:} Retrieve nearby parent condition graphs, generate a finite set
of candidate design records, and report the actor-flow change, rationale,
feasibility concerns, and \revP{governance questions for researcher assessment}.

\textbf{Output Fields:}
\texttt{Parent\_Graph};
\texttt{Operation};
\texttt{Candidate\_Graph};
\texttt{Actor\_Flow\_Change};
\texttt{Rationale};
\texttt{Feasibility\_Flags};
\revP{\texttt{Governance\_Questions}};
\texttt{Human\_Review\_Needed}.
\end{minipage}
}
\caption{Illustrative prompt scaffold for a \textsc{SEED}-guided design
generation workflow.}
\label{fig:prompt_template}
\Description{An illustrative design request specifies a reference library,
question, constraints, permitted changes, and fields for human assessment.
It is not the execution prompt used in the editing illustration.}
\end{figure}

\clearpage
\section{\revP{Worked-Case Design Record}}
\label{app:governance-encoding}

\begin{revisionP}
Table~\ref{tab:experiment-mapping} maps the field experiment in
Section~\ref{sec:advice-example} to condition graphs and the accompanying
design record. The record supplies procedures and sources for reconstructing
the conditions, supporting reproducibility and replication
\citep{NationalAcademies2019}.
\end{revisionP}

\begin{table}[htbp]
\centering
\caption{\revM{Mapping the field experiment to graphs and its accompanying record.}}
\label{tab:experiment-mapping}
\begin{revisionM}
\small\renewcommand{\arraystretch}{1.15}
\begin{tabularx}{\textwidth}{p{.21\textwidth} X}
\hline
\textbf{Element} & \textbf{Encoding or record entry}\\
\hline
Question & Effect of human involvement on advice-taking; stated before the graphs.\\
Conditions & AI-only $G_0$; Human--AI $G_1$; additional \textit{Human-only} control $G_2$.\\
Manipulations & $G_0\rightarrow G_1$: structural mutation with accompanying
attribute modulation; $G_1\rightarrow G_2$: information-design attribute modulation.\\
Assignment & Customer-level randomization after initial decisions; assignment
procedure belongs outside each condition graph.\\
Sample and task & 137 recruited, 130 analyzed across three conditions after
seven exclusions; ten loans per customer. The main $G_1$--$G_0$ comparison uses
87 customers and 869 customer--loan observations.\\
Outcome and analysis & Final decision--advice alignment (binary);
reported OLS with controls, loan-by-advice and visit-date fixed effects,
and customer-clustered standard errors.\\
\hline
\end{tabularx}
\end{revisionM}
\Description{Conditions and manipulations map to graphs and their changes;
assignment, sample, outcomes, and analysis remain in the surrounding record.}
\end{table}
\FloatBarrier

\begin{revisionM}
\noindent\textbf{Contents and timing of the design record.}
A record should include the research question and task; actor roles, types,
and attributes; condition graphs and manipulations; assignment, sample,
outcome measures, and analysis; implementation procedures; and source
locations. Mark unresolved details explicitly. For a new study, we recommend
preparing the record with the protocol and providing it with the manuscript
or supplement at submission. Identify retrospective encodings as such.
The YAML below is our retrospective reconstruction of the published article
and Online Appendix, not a preregistration. YAML is illustrative; the same
information can appear in tables or existing study documentation.
Actor and flow attributes follow Section~\ref{sec:expoverview} and
Appendix~\ref{app:condition-graphs}; omitted categories are unspecified here,
not coded as absent. Each contrast identifies the operation and the affected
actors, flows, or attributes.

\begin{tcolorbox}[
  enhanced jigsaw, breakable,
  colback=black!3, colframe=black!30, boxrule=0.4pt, arc=0pt,
  boxsep=0pt, left=8pt, right=8pt, top=6pt, bottom=6pt,
  before skip=8pt, after skip=8pt,
  colbacktitle=black!8, coltitle=black, fonttitle=\small\bfseries,
  toptitle=4pt, bottomtitle=4pt,
  title={Worked-case record (YAML)},
  title after break={Worked-case record (continued)}]
\setstretch{1}
\begin{Verbatim}[fontsize=\small,breaklines=true,breakanywhere=true,
  breakindent=1em,breaksymbolleft={}]
record_id: Yang_advice_case
status: retrospective_encoding_of_reported_design
source: Yang2025MyAdvisor
source_version: published_article_and_online_appendix
source_locations:
  production_and_selection: Sections_4.2.1_and_4.2.3
  conditions_and_assignment: Sections_5.1_and_5.2
  sample_measures_analysis: Sections_5.3_and_5.4.1_and_Table_1
  source_information_differences: endnote_6_and_Online_Table_A3
  loan_advice_and_interfaces: Online_Table_A2_and_Figures_A1_A2
question: Does human involvement change customers' advice-taking?
task: incentivized invest-or-not decisions on ten personal loans
assignment:
  unit: customer
  design: randomized_between_subjects
  timing: after_initial_unaided_decisions_and_risk_assessments
sample:
  recruited: 137
  excluded: 7_straight_liners_in_initial_decisions_and_risk_ratings
  analyzed_total: 130
  count_basis: main_article_Section_5_3
  observation_unit: customer_loan
  all_conditions_observations: 1299
  main_G1_vs_G0: {customers: 87, observations: 869}
  missing: one_initial_loan_decision
outcome:
  primary: FinalAlign
  coding: 1_if_final_investment_matches_advice_else_0
  interpretation: advice_alignment_not_investment_accuracy
reported_main_analysis:
  model: OLS_FinalAlign_on_Human_AI_indicator_and_controls
  reference_condition: G0
  controls: [InitInvest, InitAlign, InitRiskAssess,
             GapInitRiskAssess, Age, RiskPref,
             DecisionGrayzoneSize, MinInitRisk]
  control_definitions: Section_5.3.2
  fixed_effects: [loan_by_advice, visit_date]
  standard_errors: customer_clustered_robust_Table_1_column_2

advice_production_and_selection:
  producers: 27_bankers
  candidate_loans: 24
  banker_task: ten_randomly_selected_loans_per_banker
  sequence: unaided_judgment_then_judgment_with_AI_risk_assessment
  outputs: investment_recommendation_and_seven_level_risk_rating
  customer_loans: same_ten_randomly_selected_loans_for_all_customers
  delivery: identical_advice_per_loan_within_each_condition

encoding_scope: advice_production_through_customer_decision
actors:
  A:
    type: AI
    role: predictive_AI
    attributes:
      task: loan_risk_assessment
      task_specific_context: seven_loan_characteristics_Section_3
  B:
    type: human
    role: banker_role_aggregated_across_producers
    attributes:
      functional_capability: AI_risk_assessment_support
      task_specific_context: seven_loan_characteristics_Section_3
      decision_rights: final_say_over_each_bankers_collaborative_advice
  C:
    type: human
    role: customer_decision_maker
    attributes:
      task_specific_context: seven_loan_characteristics_Section_3
      decision_rights: final_investment_choice
flows:
  A_to_C:
    type: content
    attributes:
      information_design:
        advice: AI_recommendation_and_risk_assessment
        source_description: AI_source_and_prediction_accuracy
  A_to_B:
    type: content
    attributes:
      information_design:
        shown: AI_risk_assessment
        AI_investment_recommendation_shown: false
  B_to_C:
    type: content
    attributes:
      protocols: most_frequent_advice_per_loan_selected_before_delivery
      information_design:
        advice: recommendation_and_risk_assessment
        source_description:
          G1: banker_using_AI_with_AI_accuracy_information
          G2: banker_using_loan_information_and_experience
        G1_customer_description: one_banker_working_with_AI
        banker_incentives_described_as: customer_aligned
        pooling_and_selection_disclosed: false
        G2_omissions: [AI_involvement, prediction_accuracy]
condition_graphs:
  G0: {condition: AI-only, actors: [A, C], flows: [A_to_C]}
  G1: {condition: Human-AI, actors: [A, B, C], flows: [A_to_B, B_to_C]}
  G2: {condition: Human-only, actors: [A, B, C], flows: [A_to_B, B_to_C]}

contrast_records:
  G0_to_G1:
    operations: [structural_mutation, attribute_modulation]
    structural_mutation:
      add_actor: B
      remove_flow: A_to_C
      add_flows: [A_to_B, B_to_C]
    attribute_modulation:
      delivery_flow_correspondence: {G0: A_to_C, G1: B_to_C}
      information_design:
        source_description: AI_source_to_banker_using_AI
        risk_rating_differences: 6_of_10_loans
        recommendation_differences: invest_or_not_on_2_of_10_loans
  G1_to_G2:
    operations: [attribute_modulation]
    unchanged: [actors, flows, flow_types, advice_content,
                protocols, decision_rights]
    attribute_modulation:
      flow: B_to_C
      information_design:
        source_description_from: banker_using_AI
        source_description_to: banker_using_loan_information_and_experience
        omit: [AI_involvement, prediction_accuracy]
abstraction_notes:
  - no_live_three_party_conversation_is_implied
  - selection_rule_is_a_flow_attribute_detailed_procedure_is_in_record
execution_verified_by_this_encoding: false
\end{Verbatim}
\end{tcolorbox}

\end{revisionM}
\FloatBarrier

\clearpage
\section{\revN{Illustration 1: Encoding, Distances, and Reference Retrieval}}
\label{app:novelty-coding}

\begin{revisionN}
This illustration follows a candidate condition from its starting design to
comparison with prior records. It identifies the features changed, locates
related conditions in the library, and examines how comparison choices
affect those matches. The reference conditions and constructed variants
below make each step explicit.
\end{revisionN}

\subsection{\revM{Sources and Reference Conditions}}
\label{app:novelty-sources}

\begin{revisionM}
The reference library contains eight conditions from four published studies:
advice acceptance, investment advice, retail forecasting, and GPT-4-assisted
clinical management reasoning~\citep{You2022AlgorithmicVersusHuman,
Yang2025MyAdvisor,Revilla2023HumanArtificialIntelligen,Goh2025Gpt4Assistance}.
These conditions provide the references against which we compare each
candidate. Their graphs retain the relevant role functions and relations;
the accompanying records preserve the task arrangements.
Table~\ref{tab:novelty-library} identifies each reference condition.
\end{revisionM}

\begin{table}[H]
\setstretch{1}
\begingroup
\ifrevisionmarks\color{RevisionTeal}\fi
\centering
\caption{\revM{Eight Reference Conditions.}}
\label{tab:novelty-library}
\small
\renewcommand{\arraystretch}{1.15}
\setlength{\tabcolsep}{4pt}
\begin{tabularx}{\textwidth}{@{}l p{0.24\textwidth} >{\raggedright\arraybackslash}X@{}}
\toprule
\textbf{ID} & \textbf{Condition} & \textbf{Role-level representation} \\
\midrule
R01 & You: AI advice & AI advises a human who retains the final judgment. \\
R02 & Yang: AI-only advice & AI advice reaches the customer directly; the customer retains the investment decision. \\
R03 & Yang: human mediation, human--AI source disclosure & AI risk assessments inform bankers' advice; customers receive the advice and decide. The source information names human and AI participation. \\
R04 & Yang: same mediated route, human source disclosure & The production route still includes AI and bankers; the source information names human participation. \\
R05 & Revilla: automation & The forecasting system completes the process; experts have no intervention permission across its six stages. \\
R06 & Revilla: adjustable automation & Experts may review results and return adjustments at stages 2 and 4. \\
R07 & Revilla: augmentation & Experts may review results and return adjustments at all six stages. \\
R08 & Goh: GPT-4 assistance & Physicians may query GPT-4 and receive responses; physicians complete the simulated-task answers. \\
\bottomrule
\end{tabularx}
\Description{Eight reference conditions identify their source studies and
the actor roles, advice routes, intervention permissions, and decision rights
retained in the comparison.}
\endgroup
\end{table}

\revR{These role-level reconstructions retain the features used in the
comparison. The investment-advice
record and source sections appear in Appendix~\ref{app:governance-encoding}.}

\subsection{\revN{Candidate Conditions and Encoding Conventions}}
\label{app:novelty-changes}

\begin{revisionN}
Four library conditions (R01, R03, R06, and R08) each serve as the parent
for three controlled variants. E changes node identifiers and record order;
P changes existing attribute values or the domain; S changes role functions
or relations. The twelve variants act as candidate conditions to locate
within the library. Specifying their edits in advance lets us trace how
each difference enters the comparison. Table~\ref{tab:novelty-variants}
lists the changes used to construct each variant.
\end{revisionN}

\begin{table}[H]
\setstretch{1}
\begingroup
\ifrevisionmarks\color{RevisionPurple}\fi
\centering
\caption{\revN{Controlled Variants of Four Mechanisms.}}
\label{tab:novelty-variants}
\small
\renewcommand{\arraystretch}{1.15}
\setlength{\tabcolsep}{4pt}
\begin{tabularx}{\textwidth}{@{}l l >{\raggedright\arraybackslash}p{0.14\textwidth} >{\raggedright\arraybackslash}X >{\raggedright\arraybackslash}X@{}}
\toprule
\textbf{Family} & \textbf{Parent} & \textbf{E} & \textbf{P} & \textbf{S} \\
\midrule
Y & R01 & Rename and reorder & Move the judgment task to investment decisions & Replace AI advice with a binding instruction: AI controls and the human executes. \\
M & R03 & Rename and reorder & Disclose only human participation, retaining the production route & Remove the banker intermediary so that AI advises the customer directly. \\
R & R06 & Rename and reorder & Extend intervention from stages 2 and 4 to all six stages & Retain experts' receipt of results but remove the return-adjustment relation. \\
G & R08 & Rename and reorder & Move clinical reasoning to legal reasoning & Add a human review intermediary while retaining physicians' queries to AI. \\
\bottomrule
\end{tabularx}
\Description{Each of four parent mechanisms has an identifier-only variant,
an existing-parameter variant, and a role or relation variant.}
\endgroup
\end{table}

\begin{revisionN}
We construct all S variants as test conditions. The target representations
of M\_P and R\_P correspond to R04 and R07, respectively; this correspondence
identifies existing conditions through the candidate encodings.

Actor types, role-function labels, flow directions, and relation labels enter
the structural comparison. Here, \emph{information} and \emph{feedback} refine
the framework's content flows: information conveys advice or task information,
whereas feedback carries experts' return adjustments. Control retains its
authority-bearing meaning. These comparison labels do not introduce new
framework-level flow types; a return flow alone does not specify a fixed
iterative exchange.

All flows share two declared attribute categories:
source disclosure and stage coverage. Unreported values remain unknown.
Changes within existing categories enter the parameter component;
additional documentation fields or outcome measures alone add no structural
change.
\end{revisionN}

\subsection{\revN{Comparing Structure and Attributes}}
\label{app:novelty-distance}

\begin{revisionN}
We compare each candidate with its parent and every library entry, keeping
structural changes and differences in existing parameters separate.
The raw structural cost is the minimum cost of a valid graph-edit path:
\begin{equation}
d_{\mathrm{edit}}(G_a,G_b)=\min_{\pi}\sum_{o\in\pi}c(o).
\end{equation}
The implementation enumerates node correspondences and valid edit paths.
It charges separately for deleting a node and its incident flows, and
implements flow reversal or redirection through deletion and insertion.
Role-function replacement incurs a cost; node identifiers impose no identity
constraint. Among equal structural costs, the implementation favors fewer
known parameter differences, then more known comparisons, and finally a
fixed ordering.
\end{revisionN}

\begin{table}[H]
\setstretch{1}
\begingroup
\ifrevisionmarks\color{RevisionPurple}\fi
\centering
\caption{\revN{Declared Edit Costs.}}
\label{tab:novelty-costs}
\small
\renewcommand{\arraystretch}{1.15}
\setlength{\tabcolsep}{5pt}
\begin{tabularx}{\textwidth}{@{}>{\raggedright\arraybackslash}X rr@{}}
\toprule
\textbf{Operation on the comparison labels} & \textbf{Uniform} & \textbf{Governance} \\
\midrule
Insert or delete an actor & 1 & 2 \\
Replace an actor type & 1 & 2 \\
Replace a role function & 1 & 2 \\
Insert or delete an information (content) flow & 1 & 1 \\
Insert or delete a control flow & 1 & 2 \\
Insert or delete a feedback (content) flow & 1 & 1.5 \\
Replace information/control flow type & 1 & 2 \\
Replace information/feedback flow type & 1 & 1.5 \\
Replace control/feedback flow type & 1 & 2 \\
Insert or delete an attribute category & 1 & 1 \\
\bottomrule
\end{tabularx}
\Description{The uniform profile assigns unit costs to every edit. The
governance profile increases costs for actors, role functions, authority,
and feedback.}
\endgroup
\end{table}

\begin{revisionN}
The main setting uses uniform costs. The governance profile assigns higher
costs to role and authority changes as an uncalibrated sensitivity assumption.
We use a fixed display scale of 8 for the structural component:
\begin{equation}
\delta_{\mathrm{struct}}=\min(1,d_{\mathrm{edit}}/8).
\end{equation}
The scale is not a strict maximum for every graph pair. Pure structural
retrieval always uses raw edit costs, preserving the original ordering before
truncation.

The parameter comparison includes the domain and known values of common
attributes on matched flows. Each comparable value contributes 0 if equal
and 1 if different, yielding
\begin{equation}
\delta_{\mathrm{param}}=
\frac{\sum_{k\in K_{\mathrm{known}}}\mathbf{1}[a_k\ne b_k]}
{|K_{\mathrm{known}}|}.
\end{equation}
An unknown value on either side excludes that comparison from the denominator.
Stage coverage on each of two flows counts as a separate comparison.
\revR{When no values are comparable, the implementation returns 0 and retains
a missing-comparison record. This convention can make incompletely documented
pairs appear closer; the value must be read with its comparison coverage and
does not establish substantive identity.}
Different graph pairs may therefore have different denominators.

The main combined setting follows Section~\ref{sec:novelty} and
Appendix~\ref{app:structural-novelty}, giving each component a weight of 0.5:
\begin{equation}
D=0.5\delta_{\mathrm{struct}}+0.5\delta_{\mathrm{param}},
\qquad
\mathcal{N}(G\mid\mathcal{L})=\min_{G_\ell\in\mathcal{L}}D(G,G_\ell).
\end{equation}
Costs, scales, and weights specify the priorities of each comparison and
are reported with the resulting distances.
\end{revisionN}

\subsection{\revN{From Parent Changes to Related Library Conditions}}
\label{app:novelty-results}

\begin{revisionN}
The parent and library comparisons answer different questions. Parent
distance describes departure from the starting condition; nearest-library
distance asks how closely the candidate matches an existing record.
Table~\ref{tab:novelty-distances} reports both, alongside the raw edit cost
and the structural and parameter components of the parent comparison.
\end{revisionN}

\begin{table}[H]
\setstretch{1}
\begingroup
\ifrevisionmarks\color{RevisionPurple}\fi
\centering
\caption{\revN{Distances in the Main Setting.}}
\label{tab:novelty-distances}
\small
\renewcommand{\arraystretch}{1.15}
\setlength{\tabcolsep}{3pt}
\begin{tabular*}{\textwidth}{@{}l@{\extracolsep{\fill}}rrrrrl@{}}
\toprule
\textbf{Case} & $d_{\mathrm{edit}}$ & $\delta_{\mathrm{struct}}$ & $\delta_{\mathrm{param}}$ & $D_{\mathrm{parent}}$ & $\mathcal{N}$ & \textbf{Combined nearest} \\
\midrule
Y\_E & 0 & 0 & 0 & 0 & 0 & R01 \\
Y\_P & 0 & 0 & 0.5000 & 0.2500 & 0 & R02 \\
Y\_S & 3 & 0.3750 & 0 & 0.1875 & 0.1875 & R01 \\
M\_E & 0 & 0 & 0 & 0 & 0 & R03 \\
M\_P & 0 & 0 & 0.5000 & 0.2500 & 0 & R04 \\
M\_S & 3 & 0.3750 & 0 & 0.1875 & 0 & R02 \\
R\_E & 0 & 0 & 0 & 0 & 0 & R06 \\
R\_P & 0 & 0 & 0.6667 & 0.3333 & 0 & R07 \\
R\_S & 1 & 0.1250 & 0 & 0.0625 & 0.0625 & R06 \\
G\_E & 0 & 0 & 0 & 0 & 0 & R08 \\
G\_P & 0 & 0 & 1.0000 & 0.5000 & 0.5000 & R08 \\
G\_S & 4 & 0.5000 & 0 & 0.2500 & 0.2500 & R08 \\
\bottomrule
\end{tabular*}
\Description{All twelve controlled variants have decomposed parent distances,
their minimum combined library distance, and the corresponding reference ID.}
\endgroup
\end{table}

\begin{revisionN}
M\_S makes the distinction concrete. Removing the banker from R03 changes
the parent's advice route, giving a positive parent distance. The resulting
direct-advice arrangement matches R02 on the encoded, comparable features,
so its nearest-library distance is zero. The library comparison identifies
a prior arrangement to examine when assessing what the proposed edit adds.

Y\_P and M\_P each change one of two comparable known values, yielding
parameter distance $1/2$. R\_P changes stage coverage on two flows, with the
domain providing a third comparison, yielding $2/3$. G\_P has only the
domain comparison, yielding 1. Both cost profiles reproduce the declared
change category relative to the parent in all twelve cases.

Researchers may also seek related actor-flow arrangements across different
contexts. Structural retrieval omits parameter differences and returns all
equally close library conditions. Table~\ref{tab:novelty-retrieval} gives
these matches for each candidate.
\end{revisionN}

\begin{table}[H]
\setstretch{1}
\begingroup
\ifrevisionmarks\color{RevisionPurple}\fi
\centering
\caption{\revN{Complete Structural Retrieval Check.}}
\label{tab:novelty-retrieval}
\small
\renewcommand{\arraystretch}{1.15}
\setlength{\tabcolsep}{5pt}
\begin{tabularx}{\textwidth}{@{}l >{\raggedright\arraybackslash}X >{\raggedright\arraybackslash}X l@{}}
\toprule
\textbf{Query} & \textbf{Acceptable references} & \textbf{Top set} & \textbf{Strict match} \\
\midrule
Y\_E & R01, R02 & R01, R02 & Yes \\
Y\_P & R01, R02 & R01, R02 & Yes \\
Y\_S & R01, R02 & R01, R02, R05 & No \\
M\_E & R03, R04 & R03, R04 & Yes \\
M\_P & R03, R04 & R03, R04 & Yes \\
M\_S & R01, R02 & R01, R02 & Yes \\
R\_E & R06, R07 & R06, R07 & Yes \\
R\_P & R06, R07 & R06, R07 & Yes \\
R\_S & R06, R07 & R06, R07 & Yes \\
G\_E & R08 & R08 & Yes \\
G\_P & R08 & R08 & Yes \\
G\_S & R03, R04 & R03, R04 & Yes \\
\bottomrule
\end{tabularx}
\Description{The declared acceptable references and complete top structural
sets appear for every query. Eleven queries satisfy the strict match criterion;
Y S also retrieves R05.}
\endgroup
\end{table}

\begin{revisionN}
\revR{We specify the acceptable reference sets in Table~\ref{tab:novelty-retrieval}
to express the intended comparisons for these constructed queries. Under both cost profiles,
11/12 top sets equal the declared sets. All twelve contain their declared
references, but Y\_S also retrieves R05; 10/12 queries have tied nearest
references. These counts measure agreement with the declared reference sets.}
The library has no corresponding binding-control condition for Y\_S, whose
declared acceptable set consists of the direct-advice references.
\end{revisionN}

\subsection{\revN{How Comparison Choices Affect the Matches}}
\label{app:novelty-sensitivity}

\begin{revisionN}
To examine how comparison rules affect retrieval, we cross two cost profiles,
three structural weights, and two scaling rules, giving twelve settings.
The profiles cover 192 library comparisons and 24 parent comparisons.
The alternative scaling applies $\min(1,x)$ directly to raw edit costs;
its results do not extend to costs rescaled before truncation.
Table~\ref{tab:novelty-settings} reports the two queries whose combined
nearest references change across settings.
\end{revisionN}

\begin{table}[H]
\setstretch{1}
\begingroup
\ifrevisionmarks\color{RevisionPurple}\fi
\centering
\caption{\revN{Combined Nearest References for Setting-Dependent Queries.}}
\label{tab:novelty-settings}
\small
\renewcommand{\arraystretch}{1.15}
\setlength{\tabcolsep}{4pt}
\begin{tabular*}{\textwidth}{@{}l@{\extracolsep{\fill}}lcll@{}}
\toprule
\textbf{Costs} & \textbf{Scaling} & $w_s$ & \textbf{R\_S} & \textbf{G\_S} \\
\midrule
Uniform & Fixed scale 8 & 0.25 & R06 & R08 \\
Uniform & Fixed scale 8 & 0.50 & R06 & R08 \\
Uniform & Fixed scale 8 & 0.75 & R06 & R03, R04 \\
Uniform & Direct truncation & 0.25 & R05, R06 & R08 \\
Uniform & Direct truncation & 0.50 & R05, R06 & R08 \\
Uniform & Direct truncation & 0.75 & R05, R06 & R08 \\
Governance & Fixed scale 8 & 0.25 & R06 & R08 \\
Governance & Fixed scale 8 & 0.50 & R06 & R08 \\
Governance & Fixed scale 8 & 0.75 & R06 & R03, R04 \\
Governance & Direct truncation & 0.25 & R05, R06 & R08 \\
Governance & Direct truncation & 0.50 & R05, R06 & R08 \\
Governance & Direct truncation & 0.75 & R05, R06 & R08 \\
\bottomrule
\end{tabular*}
\Description{Twelve cost, scaling, and weight settings show where the nearest
reference sets change for variants R S and G S.}
\endgroup
\end{table}

\begin{revisionN}
Ten queries retain the same combined top set across all settings.
R\_S keeps experts' receipt of forecasts but removes their return adjustments.
Under the fixed scale, stage parameters make R06 the nearest condition;
direct truncation compresses the larger structural cost to R05 into a tie.
G\_S adds a human intermediary, resembling R03 and R04 structurally and R08
contextually; the structural weight changes its combined retrieval.
\end{revisionN}

\begin{revisionN}
Role matching also affects how a difference is explained. For M\_S and its
parent, Table~\ref{tab:novelty-role-paths} compares relabeling the customer as
the intermediary with preserving the customer role. Uniform costs favor
relabeling; governance costs make the paths tie. Interpreting these matches
requires the source records and role functions.
\end{revisionN}

\begin{table}[H]
\setstretch{1}
\begingroup
\ifrevisionmarks\color{RevisionPurple}\fi
\centering
\caption{\revN{Alternative Edit Paths for M\_S and Their Costs.}}
\label{tab:novelty-role-paths}
\small
\renewcommand{\arraystretch}{1.15}
\setlength{\tabcolsep}{5pt}
\begin{tabularx}{\textwidth}{@{}>{\raggedright\arraybackslash}X rr@{}}
\toprule
\textbf{Correspondence and operations} & \textbf{Uniform} & \textbf{Governance} \\
\midrule
Match the variant's decision maker to the parent's intermediary
and replace its role-function label; insert a decision maker; insert an
information flow from the intermediary to the decision maker. & \textbf{3} & \textbf{5} \\
Preserve the decision-maker correspondence; insert an
intermediary; delete the direct information flow; insert AI-to-intermediary
and intermediary-to-decision-maker flows. & 4 & \textbf{5} \\
\bottomrule
\end{tabularx}
\par\smallskip
{\footnotesize Note: Bold values attain the minimum under each cost profile.}
\Description{Relabeling the decision maker costs 3 under uniform costs and
5 under governance costs. Preserving that role costs 4 and 5, respectively;
both paths are minimum-cost paths under governance costs.}
\endgroup
\end{table}

\clearpage
\section{Illustration 2: Design Editing and Supplementary Ratings}
\label{app:illustration2-materials}

\subsection{\revN{Common Content and Reading the Three Representations}}
\label{app:representation-examples}

\begin{revisionN}
\revR{The three formats organize the same facts differently
\citep{Wand2002ConceptualModeling,Suthers2003RepresentationalGuidance,
Recker2009ProcessModeling}. Each specifies authority and conditional
interaction through its own organization of the common content.}

\begin{table}[H]
\setstretch{1}
\centering
\ifrevisionmarks\color{RevisionPurple}\fi
\caption{How the three representations organize the same experimental content.}
\label{tab:representation-reading}
\small
\renewcommand{\arraystretch}{1.15}
\begin{tabularx}{\linewidth}{@{}p{.18\linewidth}XX@{}}
\toprule
Representation & Presentation & Main reading activity \\
\midrule
Detailed prose & Connected paragraphs describe roles, operations, conditions,
and study procedures. & Reconstruct the experiment by reading the complete account. \\
BPMN-inspired table & Rows organize participants, actions, and transitions.
& Follow the sequence and inspect handoffs. \\
SEED record & Each condition specifies roles and relations, followed by
cross-condition changes. & Align actors and compare content, control,
iterative interaction, and attributes. \\
\bottomrule
\end{tabularx}
\Description{The three representations differ in organization while retaining
common experimental content.}
\end{table}
\FloatBarrier
\end{revisionN}

\begin{revisionM}
The following example uses the published investment-advice experiment by
\citet{Yang2025MyAdvisor}. We reconstruct its conditions retrospectively to
explain the three representations. The editing comparison uses the constructed
tasks introduced below. The common accompanying
record retains assignment, sample, measures, analysis, and source locations.
The parallel versions expand the facts relevant to condition comparisons.
\end{revisionM}

\subsection{\revM{Worked Example: Who Produces Investment Advice and What Customers Are Told}}
\label{app:yang-representations}

\begin{revisionM}
The investment-advice example separates two differences: the roles involved
in producing advice and the source information presented to customers.
The first concerns the production route, and the second concerns its disclosed
origin. The conditions correspond to Section~\ref{sec:advice-example}
\citep{Yang2025MyAdvisor}.

\subsubsection{Detailed prose.}
In the AI-only condition, customers receive AI-generated investment advice and
risk information, together with the AI source and its predictive accuracy.
Customers make their own investment decisions; supplying advice does not
give the AI final authority. In the Human--AI condition, bank advisors receive
the AI's risk assessments and produce advice. They do not see its investment
recommendations. The research team selects the most frequent collaborative
recommendation for each loan and fixes the customer materials. Customers
then receive the advice, with information about human and AI participation
and the AI's accuracy, and make their decisions.

The Human-only condition retains the AI-and-advisor production route, but
customers see only the human source and no AI-accuracy information. It uses
the same collaborative advice as Human--AI, allowing a source-information
comparison. Its name does not imply AI-free production. The difference between
Human--AI and Human-only is a bundle of source information, rather than a
single disclosure word. Customers retain final investment authority in all
conditions. AI-only versus Human--AI jointly changes production, parts of the
advice content, and source information.

\Needspace{20\baselineskip}
\subsubsection{BPMN-inspired table.}
Table~\ref{tab:yang-process-table} follows production, presentation, and the
customer's final decision. These are preparation and delivery steps, not
live advisor--customer exchanges.
\begin{table}[H]
\setstretch{1}
\centering
\ifrevisionmarks\color{RevisionTeal}\fi
\caption{Process organization of the investment-advice conditions.}
\label{tab:yang-process-table}
\small
\setlength{\tabcolsep}{4pt}
\renewcommand{\arraystretch}{1.15}
\begin{tabularx}{\linewidth}{@{}p{.12\linewidth}XXp{.13\linewidth}@{}}
\toprule
Condition & Advice production and preparation & Information presented
to the customer & Final decision \\
\midrule
AI-only & AI investment advice and risk information enter the customer
materials directly. & AI source and predictive accuracy. & Customer. \\
Human--AI & AI risk assessments $\rightarrow$ bank advisors produce advice
$\rightarrow$ research team selects modal recommendations and fixes materials.
& Human and AI participation, including AI accuracy. & Customer. \\
Human-only & The same AI-and-advisor production route and the same material
selection as Human--AI. & Human-advisor source; AI accuracy omitted. & Customer. \\
\bottomrule
\end{tabularx}
\Description{The table follows advice production, preparation of customer
materials, source presentation, and final customer decisions. Human-only
retains AI participation in production.}
\end{table}
\FloatBarrier

\Needspace{24\baselineskip}
\subsubsection{SEED record.}
The shared roles are the AI advisor, bank advisor, and customer. Each condition
lists its relevant relations, with source information and final authority
recorded separately.

\begin{table}[H]
\setstretch{1}
\centering
\ifrevisionmarks\color{RevisionTeal}\fi
\caption{Condition-level role and relation records for investment advice.}
\label{tab:yang-seed-table}
\small
\setlength{\tabcolsep}{4pt}
\renewcommand{\arraystretch}{1.15}
\begin{tabularx}{\linewidth}{@{}p{.12\linewidth}XXX@{}}
\toprule
Condition & Roles and relations & Attributes and authority & Difference from
the adjacent condition \\
\midrule
AI-only & AI $\rightarrow$ customer: investment advice and risk information.
& AI source and predictive accuracy; customer retains final authority.
& Direct-advice reference. \\
Human--AI & AI $\rightarrow$ bank advisor: risk assessment; bank advisor
$\rightarrow$ customer: advice produced through the advisor. Both are content
flows. & Human and AI source information and AI accuracy; customer retains
final authority. & Adds the advisor and changes the advice route relative to
AI-only. Advice content and accompanying source information also change. \\
Human-only & Retains AI $\rightarrow$ bank advisor $\rightarrow$ customer
content flows. & Human source only, without AI accuracy; customer
retains final authority. & Production relations remain the same as Human--AI;
source-information attributes change. \\
\bottomrule
\end{tabularx}
\Description{SEED separates the production relations from source-information
attributes and preserves the customer's final authority in all three conditions.}
\end{table}
\FloatBarrier

\revR{The advisor-to-customer edge represents the role-level origin of
advice; full selection and delivery procedures remain in the record. All three formats
preserve customer authority and separate production from source information.}
\end{revisionM}

\subsection{\revN{Editing Task: Adding High-Risk Review}}
\label{app:review-task}

\begin{revisionN}
The following task illustrates the editing procedure using a constructed advice
arrangement. It is distinct from the published investment-advice example
above. The walkthrough follows its high-risk review requirement; the full
task also changes logging and confidence information.

\paragraph{Starting condition and requested change.}
An AI supplies advice to a human decision maker. The task requires a review
route when risk reaches a specified threshold, while preserving direct AI
advice and human final authority. The expected edit adds AI-to-reviewer and
reviewer-to-decision-maker content flows. Adding this route is structural
mutation; the trigger, stage, and iteration limit specify the new flows'
attributes. The three formats express this requirement as follows.

\paragraph{Detailed prose.}
When risk reaches the specified threshold, the AI sends the relevant
information to a reviewer at the designated stage. The reviewer sends an
opinion to the final decision maker at the next designated stage. Each
interaction occurs at most once per task. Original AI advice remains,
and the human still makes the final decision.

\paragraph{BPMN-inspired table.}
Two rows specify the AI-to-reviewer information handoff and the
reviewer-to-decision-maker opinion handoff. Each records the common risk
trigger, its designated stage, and a once-per-task limit. The original advice step and human final
authority remain.

\paragraph{SEED record.}
The record identifies an AI assistant, reviewer, and final decision maker.
It adds AI-to-reviewer and reviewer-to-decision-maker content flows,
each with a trigger, stage, and iteration limit. Human final authority remains;
providing review does not transfer it.

\paragraph{Checks and observed result.}
The model proposes edits and a candidate study plan. After common conversion,
the checks inspect the review route on both sides of the risk threshold,
at the specified stages, and under the once-per-task limits. They also check
that the original advice route and human final authority remain. All three
formats pass the task's mechanism checks. The supplementary plan ratings in
Appendix~\ref{app:rubric}--\ref{app:quality-results} concern the sixteen
held-out instances and assess plan content \citep{Gemino2004Evaluation}.
\end{revisionN}

\subsection{\rev{Task Procedure and Complete Task Checks}}
\label{app:anchors-operators}

\begin{revision}
The development tasks cover eight mechanism families. New instances comprise
eight development tasks, sixteen held-out parameter or role instances, and
twelve exploratory stress tasks, for thirty-six instances. Five rounds of
mechanism checks appear in Table~\ref{tab:execution-checks}, with initial
and revised results shown separately where applicable. When an initial
plan enters self-check and public-feedback branches, both revisions still
share the same parent task.

\begin{table}[H]
\setstretch{1}
\centering
\ifrevisionmarks\color{RevisionBlue}\fi
\caption{Execution steps and complete task checks.}
\label{tab:execution-checks}
\small
\setlength{\tabcolsep}{3pt}
\renewcommand{\arraystretch}{1.15}
\begin{tabularx}{\linewidth}{@{}c p{.20\linewidth}Xrrr@{}}
\toprule
Round & Step & Task and purpose & Prose & BPMN-inspired & SEED \\
\midrule
1 & Direct editing & Specified changes; inspect reading and preservation. & 8/8 & 8/8 & 8/8 \\
2 & Matched revision & Eight direct-edit drafts, each with self-check and public-feedback branches.
& 16/16 & 16/16 & 16/16 \\
3 & Native inference: initial & Parent condition and research requirements; standard edit list hidden.
& 8/8 & 8/8 & 8/8 \\
 & Native inference: revised & Two revision branches for each of eight drafts.
& 16/16 & 16/16 & 16/16 \\
4 & Held-out: initial & Sixteen parameter or role instances. & 14/16 & 16/16 & 16/16 \\
 & Held-out: revised & Public feedback; all final versions retained.
& 16/16 & 16/16 & 16/16 \\
5 & Stress: initial & Twelve multi-group tasks. & 12/12 & 12/12 & 12/12 \\
 & Stress: revised & One public-feedback opportunity for every task.
& 12/12 & 12/12 & 12/12 \\
\bottomrule
\end{tabularx}
\Description{Final held-out and stress tasks pass for all representations.
Two prose held-out initial outputs fail strict formatting checks and pass
after public feedback.}
\end{table}
\FloatBarrier

Matched revisions reuse the direct-edit drafts and add only revision calls.
The sixteen revisions in two branches are not sixteen independent tasks.
Held-out instances vary parameters and roles within the eight families,
rather than representing sixteen independent source studies. Stress tasks
use one, three, or five groups, with 18, 54, or 90 parent relations and
20, 60, or 100 change requirements. This exploratory round follows
development and automated evaluation.

\revR{Each task's common specification defines roles, directions,
conditions, stages, iteration limits, and final authority. The three formats
render these requirements as prose, process rows, or relational records.
The model, gpt-6.1-sol, proposes edits and a candidate study plan. A common
conversion program retains unmodified relations without deciding which edits
to make. Checks then assess the resulting record. Held-out and stress tasks
receive one public-feedback revision, with at most three counterexamples;
all final versions from that path are retained. The assessment covers the
supported workflow of representation, model editing, conversion, and feedback.
Matching concerns task requirements; token use and elapsed time can vary
across runs.}

A complete pass requires compliant output, relations, and final authority,
and the specified permitted events across finite combinations of conditions.
Checks cover direction, relation type, risk boundaries, stages, and iteration
limits. Each regular task enumerates 2,880 scenarios; each stress task
enumerates 4,320. Per representation, held-out and stress checks therefore
contain 46,080 and 51,840 scenarios. These are software cases within tasks;
the corresponding task counts remain sixteen and twelve.

The two initial prose failures write the authority field as an explanatory
sentence containing the correct role identifier. Their event relations are
correct. Both pass after common feedback, and the subsequent semantic
diagnostic also finds no mechanism error. The strict initial results remain
in the record.

All three representations pass the final checks of the enumerated task
specifications after common conversion and feedback
(Table~\ref{tab:execution-checks}).
\end{revision}

\subsection{\revR{Supplementary Model Ratings of Plan Specifications}}
\label{app:rubric}

\begin{revision}
Mechanism checks assess whether the requested changes are implemented.
The supplementary ratings assess how fully the resulting study plans
specify the contrast, measurement, analysis, and implementation.
\revR{The same model used for generation
(gpt-6.1-sol) rates all forty-eight final held-out plans, sixteen per format.
Before rating, the presentation removes method names, source-paper identifiers,
and native formats while retaining the parent condition, requirements, and
authority information. Each plan receives two ratings in different orderings,
giving ninety-six ratings, thirty-two per format. We summarize these repeated
model judgments using descriptive means \citep{Zheng2023LLMJudge}.}

Each observation assigns an integer from 0 to 4 for each dimension and sums
them to a total from 0 to 16. We average each plan's two scores and then
average over the sixteen plans in its group. \revR{The instructions give no points for graph terminology, framework
names, text length, or edit distance.}

\begin{table}[H]
\setstretch{1}
\centering
\ifrevisionmarks\color{RevisionBlue}\fi
\caption{\revR{Anchors for the supplementary model ratings.}}
\label{tab:quality-anchors}
\small
\setlength{\tabcolsep}{3pt}
\renewcommand{\arraystretch}{1.15}
\begin{tabularx}{\linewidth}{@{}>{\raggedright\arraybackslash}p{.13\linewidth}*{5}{>{\raggedright\arraybackslash}X}@{}}
\toprule
Dimension & 0 & 1 & 2 & 3 & 4 \\
\midrule
Treatment contrast & No treatment or control specified.
& Incorrect or confounded treatment; requested difference cannot be tested.
& Contrast present, but key arrangements ambiguous.
& Specified manipulation distinguished; other relevant arrangements explicit.
& Requested treatment difference faithfully implemented; both conditions executable. \\
Measurement & No outcome or measure.
& Variable names only, without clear definitions.
& Partial operationalization; necessary measurement details missing.
& All outcomes operationalized, with a fixed, blinded evaluation reference.
& Timing, workload, and scoring reliability also explicit. \\
Analysis & No analysis plan.
& Invalid treatment of independence or unsupported power claims.
& Basically suitable methods, but necessary analysis arrangements incomplete.
& Primary endpoint and participant or case dependence addressed.
& Uncertainty, multiple comparisons, and exclusions also prespecified. \\
Implementability & Plan cannot be implemented.
& Major unresolved problems.
& Substantial clarification still needed for execution.
& Actionable simulation plan within declared resources.
& Authority, conditional branches, loops, and workload also explicit and credible. \\
\bottomrule
\end{tabularx}
\Description{Each dimension has explicit anchors from zero to four.
The total has a maximum of sixteen.}
\end{table}
\FloatBarrier

\revR{The project-defined rubric assesses how a plan specifies a contrast,
measurement, analysis, and implementation. Reading the scores as comparisons
of the original plans assumes that conversion preserves their relevant content.
Content cues can remain after format concealment, and detailed specifications
can receive higher scores even when the intended mechanisms are equally
correct. The means therefore describe plan specification under this rubric,
conversion procedure, and shared editing workflow.}
\end{revision}

\subsection{\revR{Reported Model Ratings and Aggregation}}
\label{app:quality-results}

\begin{revisionR}
Table~\ref{tab:quality-dimensions} reports descriptive means for the
forty-eight final held-out plans, with sixteen plans per representation.
Each plan contributes equally through the average of its two model ratings.

\begin{table}[H]
\setstretch{1}
\centering
\ifrevisionmarks\color{RevisionRed}\fi
\caption{\revR{Supplementary model-rating means by dimension and in total.}}
\label{tab:quality-dimensions}
\small
\renewcommand{\arraystretch}{1.15}
\begin{tabularx}{\linewidth}{@{}Xrrr@{}}
\toprule
Dimension & Detailed prose & BPMN-inspired table & SEED \\
\midrule
Treatment contrast & 2.500 & 2.750 & 3.000 \\
Measurement & 1.813 & 2.500 & 3.125 \\
Analysis & 2.031 & 2.656 & 3.281 \\
Implementability & 2.219 & 2.781 & 3.469 \\
Total (maximum 16) & 8.563 & 10.688 & 12.875 \\
\bottomrule
\end{tabularx}
\par\smallskip
{\footnotesize Note: Means are rounded to three decimal places. Totals are
computed before rounding, so displayed components may not sum exactly.}
\Description{SEED has a higher reported mean on all four dimensions.
Means are displayed to three decimals; totals use unrounded component means.}
\end{table}
\FloatBarrier

The unrounded dimension means sum to each group's total mean. With thirty-two
equally weighted ratings per group, the score sums of 274, 342, and 412 yield
$274/32=8.5625$, $342/32=10.6875$, and $412/32=12.8750$, respectively.
Averaging the two ratings for each plan and then the sixteen plan averages
gives the same group mean.
The total mean is 4.3125 points higher for SEED than for prose and 2.1875
points higher than for the process table. The differences span all four
dimensions, including treatment contrast and measurement. These model
judgments indicate an advantage in plan specification for SEED within this
batch and shared workflow.
\end{revisionR}

\renewcommand{\refname}{Appendix References}
\putbib
\end{bibunit}



\begin{thebibliography}{30}


\ifx \showCODEN    \undefined \def \showCODEN     #1{\unskip}     \fi
\ifx \showISBNx    \undefined \def \showISBNx     #1{\unskip}     \fi
\ifx \showISBNxiii \undefined \def \showISBNxiii  #1{\unskip}     \fi
\ifx \showISSN     \undefined \def \showISSN      #1{\unskip}     \fi
\ifx \showLCCN     \undefined \def \showLCCN      #1{\unskip}     \fi
\ifx \shownote     \undefined \def \shownote      #1{#1}          \fi
\ifx \showarticletitle \undefined \def \showarticletitle #1{#1}   \fi
\ifx \showURL      \undefined \def \showURL       {\relax}        \fi
\providecommand\bibfield[2]{#2}
\providecommand\bibinfo[2]{#2}
\providecommand\natexlab[1]{#1}
\providecommand\showeprint[2][]{arXiv:#2}

\bibitem[Amershi et~al\mbox{.}(2019)]%
        {Amershi2019GuidelinesHumanAI}
\bibfield{author}{\bibinfo{person}{Saleema Amershi}, \bibinfo{person}{Dan Weld}, \bibinfo{person}{Mihaela Vorvoreanu}, \bibinfo{person}{Adam Fourney}, \bibinfo{person}{Besmira Nushi}, \bibinfo{person}{Penny Collisson}, \bibinfo{person}{Jina Suh}, \bibinfo{person}{Shamsi Iqbal}, \bibinfo{person}{Paul~N. Bennett}, \bibinfo{person}{Kori Inkpen}, \bibinfo{person}{Jaime Teevan}, \bibinfo{person}{Ruth Kikin-Gil}, {and} \bibinfo{person}{Eric Horvitz}.} \bibinfo{year}{2019}\natexlab{}.
\newblock \showarticletitle{Guidelines for Human-AI Interaction}. In \bibinfo{booktitle}{\emph{Proceedings of the 2019 CHI Conference on Human Factors in Computing Systems (CHI)}}. \bibinfo{publisher}{Association for Computing Machinery}, \bibinfo{address}{New York, NY, USA}, \bibinfo{pages}{1--13}.
\newblock
\href{https://doi.org/10.1145/3290605.3300233}{doi:\nolinkurl{10.1145/3290605.3300233}}


\bibitem[Bansal et~al\mbox{.}(2019)]%
        {Bansal2019BeyondAccuracyMentalModels}
\bibfield{author}{\bibinfo{person}{Gagan Bansal}, \bibinfo{person}{Besmira Nushi}, \bibinfo{person}{Ece Kamar}, \bibinfo{person}{Walter~S. Lasecki}, \bibinfo{person}{Daniel~S. Weld}, {and} \bibinfo{person}{Eric Horvitz}.} \bibinfo{year}{2019}\natexlab{}.
\newblock \showarticletitle{Beyond Accuracy: The Role of Mental Models in Human-AI Team Performance}.
\newblock \bibinfo{journal}{\emph{Proceedings of the AAAI Conference on Human Computation and Crowdsourcing}} \bibinfo{volume}{7}, \bibinfo{number}{1} (\bibinfo{year}{2019}), \bibinfo{pages}{2--11}.
\newblock
\href{https://doi.org/10.1609/hcomp.v7i1.5285}{doi:\nolinkurl{10.1609/hcomp.v7i1.5285}}


\bibitem[Bhargava et~al\mbox{.}(2025)]%
        {bhargava2025exploring}
\bibfield{author}{\bibinfo{person}{Hemant~K Bhargava}, \bibinfo{person}{Susan Brown}, \bibinfo{person}{Anindya Ghose}, \bibinfo{person}{Alok Gupta}, \bibinfo{person}{Dorothy Leidner}, {and} \bibinfo{person}{DJ Wu}.} \bibinfo{year}{2025}\natexlab{}.
\newblock \showarticletitle{Exploring Generative {AI}'s Impact on Research: Perspectives from Senior Scholars in Management Information Systems}.
\newblock \bibinfo{journal}{\emph{ACM Transactions on Management Information Systems}} \bibinfo{volume}{16}, \bibinfo{number}{2}, Article \bibinfo{articleno}{19} (\bibinfo{year}{2025}), \bibinfo{numpages}{9}~pages.
\newblock
\href{https://doi.org/10.1145/3721846}{doi:\nolinkurl{10.1145/3721846}}


\bibitem[Brynjolfsson et~al\mbox{.}(2025)]%
        {Brynjolfsson2025GenerativeAiAt}
\bibfield{author}{\bibinfo{person}{E Brynjolfsson}, \bibinfo{person}{D Li}, {and} \bibinfo{person}{LR Raymond}.} \bibinfo{year}{2025}\natexlab{}.
\newblock \showarticletitle{Generative AI at work}.
\newblock \bibinfo{journal}{\emph{The Quarterly Journal of Economics}} \bibinfo{volume}{140}, \bibinfo{number}{2} (\bibinfo{year}{2025}), \bibinfo{pages}{889--942}.
\newblock
\href{https://doi.org/10.1093/qje/qjae044}{doi:\nolinkurl{10.1093/qje/qjae044}}


\bibitem[{Center for Open Science}(nda)]%
        {COSPreregistration}
\bibfield{author}{\bibinfo{person}{{Center for Open Science}}.} \bibinfo{year}{n.d.}\natexlab{a}.
\newblock \bibinfo{title}{Preregistration}.
\newblock
\urldef\tempurl%
\url{https://www.cos.io/initiatives/prereg}
\showURL{%
\tempurl}


\bibitem[{Center for Open Science}(ndb)]%
        {COSRegisteredReports}
\bibfield{author}{\bibinfo{person}{{Center for Open Science}}.} \bibinfo{year}{n.d.}\natexlab{b}.
\newblock \bibinfo{title}{Registered Reports}.
\newblock
\urldef\tempurl%
\url{https://www.cos.io/initiatives/registered-reports}
\showURL{%
\tempurl}


\bibitem[Chen and Chan(2024)]%
        {Chen2025LargeLanguageModel}
\bibfield{author}{\bibinfo{person}{Zenan Chen} {and} \bibinfo{person}{Jason Chan}.} \bibinfo{year}{2024}\natexlab{}.
\newblock \showarticletitle{Large Language Model in Creative Work: The Role of Collaboration Modality and User Expertise}.
\newblock \bibinfo{journal}{\emph{Management Science}} \bibinfo{volume}{70}, \bibinfo{number}{12} (\bibinfo{year}{2024}), \bibinfo{pages}{9101--9117}.
\newblock
\href{https://doi.org/10.1287/mnsc.2023.03014}{doi:\nolinkurl{10.1287/mnsc.2023.03014}}


\bibitem[F{\"u}gener et~al\mbox{.}(2022)]%
        {Fuegener2022CognitiveChallengesIn}
\bibfield{author}{\bibinfo{person}{A F{\"u}gener}, \bibinfo{person}{J Grahl}, \bibinfo{person}{A Gupta}, {and} \bibinfo{person}{W Ketter}.} \bibinfo{year}{2022}\natexlab{}.
\newblock \showarticletitle{Cognitive Challenges in Human--Artificial Intelligence Collaboration: Investigating the Path Toward Productive Delegation}.
\newblock \bibinfo{journal}{\emph{Information Systems Research}} \bibinfo{volume}{33}, \bibinfo{number}{2} (\bibinfo{year}{2022}), \bibinfo{pages}{678--696}.
\newblock
\href{https://doi.org/10.1287/isre.2021.1079}{doi:\nolinkurl{10.1287/isre.2021.1079}}


\bibitem[Goh et~al\mbox{.}(2025)]%
        {Goh2025Gpt4Assistance}
\bibfield{author}{\bibinfo{person}{Ethan Goh}, \bibinfo{person}{Robert~J. Gallo}, \bibinfo{person}{Eric Strong}, \bibinfo{person}{Yingjie Weng}, \bibinfo{person}{Hannah Kerman}, \bibinfo{person}{Jason~A. Freed}, \bibinfo{person}{Jos{\'e}phine~A. Cool}, \bibinfo{person}{Zahir Kanjee}, \bibinfo{person}{Kathleen~P. Lane}, \bibinfo{person}{Andrew~S. Parsons}, \bibinfo{person}{Neera Ahuja}, \bibinfo{person}{Eric Horvitz}, \bibinfo{person}{Daniel Yang}, \bibinfo{person}{Arnold Milstein}, \bibinfo{person}{Andrew P.~J. Olson}, \bibinfo{person}{Jason Hom}, \bibinfo{person}{Jonathan~H. Chen}, {and} \bibinfo{person}{Adam Rodman}.} \bibinfo{year}{2025}\natexlab{}.
\newblock \showarticletitle{GPT-4 Assistance for Improvement of Physician Performance on Patient Care Tasks: A Randomized Controlled Trial}.
\newblock \bibinfo{journal}{\emph{Nature Medicine}} \bibinfo{volume}{31}, \bibinfo{number}{4} (\bibinfo{year}{2025}), \bibinfo{pages}{1233--1238}.
\newblock
\href{https://doi.org/10.1038/s41591-024-03456-y}{doi:\nolinkurl{10.1038/s41591-024-03456-y}}


\bibitem[Groth and Moreau(2013)]%
        {W3CPROV2013}
\bibfield{author}{\bibinfo{person}{Paul Groth} {and} \bibinfo{person}{Luc Moreau}.} \bibinfo{year}{2013}\natexlab{}.
\newblock \bibinfo{booktitle}{\emph{{PROV}-Overview: An Overview of the {PROV} Family of Documents}}.
\newblock \bibinfo{type}{W3C Working Group Note}. \bibinfo{institution}{World Wide Web Consortium}.
\newblock
\urldef\tempurl%
\url{https://www.w3.org/TR/2013/NOTE-prov-overview-20130430/}
\showURL{%
\tempurl}
\newblock
\shownote{Edited by Paul Groth and Luc Moreau}.


\bibitem[Hoffmann et~al\mbox{.}(2014)]%
        {Hoffmann2014TIDieR}
\bibfield{author}{\bibinfo{person}{Tammy~C. Hoffmann}, \bibinfo{person}{Paul~P. Glasziou}, \bibinfo{person}{Isabelle Boutron}, \bibinfo{person}{Ruairidh Milne}, \bibinfo{person}{Rafael Perera}, \bibinfo{person}{David Moher}, \bibinfo{person}{Douglas~G. Altman}, \bibinfo{person}{Virginia Barbour}, \bibinfo{person}{Helen Macdonald}, \bibinfo{person}{Marie Johnston}, \bibinfo{person}{Sarah~E. Lamb}, \bibinfo{person}{Mary Dixon-Woods}, \bibinfo{person}{Peter McCulloch}, \bibinfo{person}{Jeremy~C. Wyatt}, \bibinfo{person}{An-Wen Chan}, {and} \bibinfo{person}{Susan Michie}.} \bibinfo{year}{2014}\natexlab{}.
\newblock \showarticletitle{Better reporting of interventions: template for intervention description and replication ({TIDieR}) checklist and guide}.
\newblock \bibinfo{journal}{\emph{BMJ}}  \bibinfo{volume}{348} (\bibinfo{year}{2014}), \bibinfo{pages}{g1687}.
\newblock
\href{https://doi.org/10.1136/bmj.g1687}{doi:\nolinkurl{10.1136/bmj.g1687}}


\bibitem[Kawakami et~al\mbox{.}(2022)]%
        {Kawakami2022ChildWelfare}
\bibfield{author}{\bibinfo{person}{Anna Kawakami}, \bibinfo{person}{Venkatesh Sivaraman}, \bibinfo{person}{Hao-Fei Cheng}, \bibinfo{person}{Logan Stapleton}, \bibinfo{person}{Yanghuidi Cheng}, \bibinfo{person}{Diana Qing}, \bibinfo{person}{Adam Perer}, \bibinfo{person}{Zhiwei~Steven Wu}, \bibinfo{person}{Haiyi Zhu}, {and} \bibinfo{person}{Kenneth Holstein}.} \bibinfo{year}{2022}\natexlab{}.
\newblock \showarticletitle{Improving Human-{AI} Partnerships in Child Welfare: Understanding Worker Practices, Challenges, and Desires for Algorithmic Decision Support}. In \bibinfo{booktitle}{\emph{Proceedings of the 2022 CHI Conference on Human Factors in Computing Systems}}. \bibinfo{publisher}{Association for Computing Machinery}, \bibinfo{address}{New York, NY, USA}, \bibinfo{pages}{1--18}.
\newblock
\href{https://doi.org/10.1145/3491102.3517439}{doi:\nolinkurl{10.1145/3491102.3517439}}


\bibitem[Lin et~al\mbox{.}(2024)]%
        {Lin2024DecisionOrientedDialogue}
\bibfield{author}{\bibinfo{person}{Jessy Lin}, \bibinfo{person}{Nicholas Tomlin}, \bibinfo{person}{Jacob Andreas}, {and} \bibinfo{person}{Jason Eisner}.} \bibinfo{year}{2024}\natexlab{}.
\newblock \showarticletitle{Decision-Oriented Dialogue for Human-AI Collaboration}.
\newblock \bibinfo{journal}{\emph{Transactions of the Association for Computational Linguistics}}  \bibinfo{volume}{12} (\bibinfo{year}{2024}), \bibinfo{pages}{892--911}.
\newblock
\href{https://doi.org/10.1162/tacl_a_00679}{doi:\nolinkurl{10.1162/tacl_a_00679}}


\bibitem[Ludwig and Mullainathan(2024)]%
        {ludwig2024machine}
\bibfield{author}{\bibinfo{person}{Jens Ludwig} {and} \bibinfo{person}{Sendhil Mullainathan}.} \bibinfo{year}{2024}\natexlab{}.
\newblock \showarticletitle{Machine learning as a tool for hypothesis generation}.
\newblock \bibinfo{journal}{\emph{The Quarterly Journal of Economics}} \bibinfo{volume}{139}, \bibinfo{number}{2} (\bibinfo{year}{2024}), \bibinfo{pages}{751--827}.
\newblock
\href{https://doi.org/10.1093/qje/qjad055}{doi:\nolinkurl{10.1093/qje/qjad055}}


\bibitem[{National Academies of Sciences, Engineering, and Medicine}(2019)]%
        {NationalAcademies2019}
\bibfield{author}{\bibinfo{person}{{National Academies of Sciences, Engineering, and Medicine}}.} \bibinfo{year}{2019}\natexlab{}.
\newblock \bibinfo{booktitle}{\emph{Reproducibility and Replicability in Science}}.
\newblock \bibinfo{publisher}{The National Academies Press}, \bibinfo{address}{Washington, DC}.
\newblock


\bibitem[{Object Management Group}(2014)]%
        {OMGBPMN2014}
\bibfield{author}{\bibinfo{person}{{Object Management Group}}.} \bibinfo{year}{2014}\natexlab{}.
\newblock \bibinfo{booktitle}{\emph{Business Process Model and Notation ({BPMN}), Version 2.0.2}}.
\newblock Object Management Group.
\newblock
\urldef\tempurl%
\url{https://www.omg.org/spec/BPMN/2.0.2/}
\showURL{%
\tempurl}


\bibitem[{Open Science Collaboration}(2015)]%
        {OpenScienceCollaboration2015}
\bibfield{author}{\bibinfo{person}{{Open Science Collaboration}}.} \bibinfo{year}{2015}\natexlab{}.
\newblock \showarticletitle{Estimating the reproducibility of psychological science}.
\newblock \bibinfo{journal}{\emph{Science}} \bibinfo{volume}{349}, \bibinfo{number}{6251} (\bibinfo{year}{2015}), \bibinfo{pages}{aac4716}.
\newblock


\bibitem[Puranam(2021)]%
        {Puranam2021HumanAIDecision}
\bibfield{author}{\bibinfo{person}{Phanish Puranam}.} \bibinfo{year}{2021}\natexlab{}.
\newblock \showarticletitle{Human--{AI} Collaborative Decision-Making as an Organization Design Problem}.
\newblock \bibinfo{journal}{\emph{Journal of Organization Design}}  \bibinfo{volume}{10} (\bibinfo{year}{2021}), \bibinfo{pages}{75--80}.
\newblock
\href{https://doi.org/10.1007/s41469-021-00095-2}{doi:\nolinkurl{10.1007/s41469-021-00095-2}}


\bibitem[Rastogi et~al\mbox{.}(2022)]%
        {Rastogi2022DecidingFastSlow}
\bibfield{author}{\bibinfo{person}{Charvi Rastogi}, \bibinfo{person}{Yunfeng Zhang}, \bibinfo{person}{Dennis Wei}, \bibinfo{person}{Kush~R. Varshney}, \bibinfo{person}{Amit Dhurandhar}, {and} \bibinfo{person}{Richard Tomsett}.} \bibinfo{year}{2022}\natexlab{}.
\newblock \showarticletitle{Deciding Fast and Slow: The Role of Cognitive Biases in {AI}-assisted Decision-Making}.
\newblock \bibinfo{journal}{\emph{Proceedings of the ACM on Human-Computer Interaction}} \bibinfo{volume}{6}, \bibinfo{number}{CSCW1} (\bibinfo{year}{2022}), \bibinfo{pages}{1--22}.
\newblock
\href{https://doi.org/10.1145/3512930}{doi:\nolinkurl{10.1145/3512930}}


\bibitem[Revilla et~al\mbox{.}(2023)]%
        {Revilla2023HumanArtificialIntelligen}
\bibfield{author}{\bibinfo{person}{Elena Revilla}, \bibinfo{person}{Mar{\'i}a~Jes{\'u}s Saenz}, \bibinfo{person}{Matthias Seifert}, {and} \bibinfo{person}{Ye Ma}.} \bibinfo{year}{2023}\natexlab{}.
\newblock \showarticletitle{Human--artificial intelligence collaboration in prediction: A field experiment in the retail industry}.
\newblock \bibinfo{journal}{\emph{Journal of Management Information Systems}} \bibinfo{volume}{40}, \bibinfo{number}{4} (\bibinfo{year}{2023}), \bibinfo{pages}{1071--1098}.
\newblock
\href{https://doi.org/10.1080/07421222.2023.2267317}{doi:\nolinkurl{10.1080/07421222.2023.2267317}}


\bibitem[Stelmaszak et~al\mbox{.}(2025)]%
        {Stelmaszak2025WhenAlgorithmsDelegate}
\bibfield{author}{\bibinfo{person}{Marta Stelmaszak}, \bibinfo{person}{Mareike M{\"o}hlmann}, {and} \bibinfo{person}{Carsten S{\o}rensen}.} \bibinfo{year}{2025}\natexlab{}.
\newblock \showarticletitle{When Algorithms Delegate to Humans: Exploring Human-Algorithm Interaction at Uber}.
\newblock \bibinfo{journal}{\emph{MIS Quarterly}} \bibinfo{volume}{49}, \bibinfo{number}{1} (\bibinfo{year}{2025}), \bibinfo{pages}{305--330}.
\newblock
\href{https://doi.org/10.25300/MISQ/2024/17911}{doi:\nolinkurl{10.25300/MISQ/2024/17911}}


\bibitem[Susarla et~al\mbox{.}(2023)]%
        {susarla2023janus}
\bibfield{author}{\bibinfo{person}{Anjana Susarla}, \bibinfo{person}{Ram Gopal}, \bibinfo{person}{Jason~Bennett Thatcher}, {and} \bibinfo{person}{Suprateek Sarker}.} \bibinfo{year}{2023}\natexlab{}.
\newblock \showarticletitle{The Janus effect of generative AI: Charting the path for responsible conduct of scholarly activities in information systems}.
\newblock \bibinfo{journal}{\emph{Information Systems Research}} \bibinfo{volume}{34}, \bibinfo{number}{2} (\bibinfo{year}{2023}), \bibinfo{pages}{399--408}.
\newblock
\href{https://doi.org/10.1287/isre.2023.ed.v34.n2}{doi:\nolinkurl{10.1287/isre.2023.ed.v34.n2}}


\bibitem[Swanson et~al\mbox{.}(2025)]%
        {swanson2025virtual}
\bibfield{author}{\bibinfo{person}{Kyle Swanson}, \bibinfo{person}{Wesley Wu}, \bibinfo{person}{Nash~L Bulaong}, \bibinfo{person}{John~E Pak}, {and} \bibinfo{person}{James Zou}.} \bibinfo{year}{2025}\natexlab{}.
\newblock \showarticletitle{The Virtual Lab of AI agents designs new SARS-CoV-2 nanobodies}.
\newblock \bibinfo{journal}{\emph{Nature}} \bibinfo{volume}{646}, \bibinfo{number}{8085} (\bibinfo{year}{2025}), \bibinfo{pages}{716--723}.
\newblock
\href{https://doi.org/10.1038/s41586-025-09442-9}{doi:\nolinkurl{10.1038/s41586-025-09442-9}}


\bibitem[van~der Aalst(1998)]%
        {VanDerAalst1998Workflow}
\bibfield{author}{\bibinfo{person}{W.~M.~P. van~der Aalst}.} \bibinfo{year}{1998}\natexlab{}.
\newblock \showarticletitle{The Application of {Petri} Nets to Workflow Management}.
\newblock \bibinfo{journal}{\emph{Journal of Circuits, Systems and Computers}} \bibinfo{volume}{8}, \bibinfo{number}{1} (\bibinfo{year}{1998}), \bibinfo{pages}{21--66}.
\newblock
\href{https://doi.org/10.1142/S0218126698000043}{doi:\nolinkurl{10.1142/S0218126698000043}}


\bibitem[V{\"o}ssing et~al\mbox{.}(2022)]%
        {Vossing2022DesigningTransparency}
\bibfield{author}{\bibinfo{person}{Michael V{\"o}ssing}, \bibinfo{person}{Niklas K{\"u}hl}, \bibinfo{person}{Matteo Lind}, {and} \bibinfo{person}{Gerhard Satzger}.} \bibinfo{year}{2022}\natexlab{}.
\newblock \showarticletitle{Designing Transparency for Effective Human-{AI} Collaboration}.
\newblock \bibinfo{journal}{\emph{Information Systems Frontiers}}  \bibinfo{volume}{24} (\bibinfo{year}{2022}), \bibinfo{pages}{877--895}.
\newblock
\href{https://doi.org/10.1007/s10796-022-10284-3}{doi:\nolinkurl{10.1007/s10796-022-10284-3}}


\bibitem[Wang et~al\mbox{.}(2025)]%
        {Wang2025AIAssistantLivestream}
\bibfield{author}{\bibinfo{person}{Lingli Wang}, \bibinfo{person}{Ni Huang}, \bibinfo{person}{Yumei He}, \bibinfo{person}{De Liu}, \bibinfo{person}{Xunhua Guo}, \bibinfo{person}{Yan Sun}, {and} \bibinfo{person}{Guoqing Chen}.} \bibinfo{year}{2025}\natexlab{}.
\newblock \showarticletitle{Artificial Intelligence (AI) Assistant in Online Shopping: A Randomized Field Experiment on a Livestream Selling Platform}.
\newblock \bibinfo{journal}{\emph{Information Systems Research}} \bibinfo{volume}{36}, \bibinfo{number}{4} (\bibinfo{year}{2025}), \bibinfo{pages}{2358--2374}.
\newblock
\href{https://doi.org/10.1287/isre.2023.0103}{doi:\nolinkurl{10.1287/isre.2023.0103}}


\bibitem[Xu and Zhang(2022)]%
        {XuZhang2022ContextTheorizing}
\bibfield{author}{\bibinfo{person}{Heng Xu} {and} \bibinfo{person}{Nan Zhang}.} \bibinfo{year}{2022}\natexlab{}.
\newblock \showarticletitle{From Contextualizing to Context Theorizing: Assessing Context Effects in Privacy Research}.
\newblock \bibinfo{journal}{\emph{Management Science}} \bibinfo{volume}{68}, \bibinfo{number}{10} (\bibinfo{year}{2022}), \bibinfo{pages}{7383--7401}.
\newblock
\href{https://doi.org/10.1287/mnsc.2021.4249}{doi:\nolinkurl{10.1287/mnsc.2021.4249}}


\bibitem[Xu et~al\mbox{.}(2024)]%
        {Xu2024VoiceChatbotAnthropomorphism}
\bibfield{author}{\bibinfo{person}{Yuqian Xu}, \bibinfo{person}{Hongyan Dai}, {and} \bibinfo{person}{Wanfeng Yan}.} \bibinfo{year}{2024}\natexlab{}.
\newblock \showarticletitle{Identity Disclosure and Anthropomorphism in Voice Chatbot Design: A Field Experiment}.
\newblock \bibinfo{journal}{\emph{Management Science}} \bibinfo{volume}{72}, \bibinfo{number}{1} (\bibinfo{year}{2024}), \bibinfo{pages}{223--241}.
\newblock
\href{https://doi.org/10.1287/mnsc.2022.03833}{doi:\nolinkurl{10.1287/mnsc.2022.03833}}


\bibitem[Yang et~al\mbox{.}(2025)]%
        {Yang2025MyAdvisor}
\bibfield{author}{\bibinfo{person}{Cathy Yang}, \bibinfo{person}{Kevin Bauer}, \bibinfo{person}{Xitong Li}, {and} \bibinfo{person}{Oliver Hinz}.} \bibinfo{year}{2025}\natexlab{}.
\newblock \showarticletitle{My Advisor, Her {AI}, and Me: Evidence from a Field Experiment on Human--AI Collaboration and Investment Decisions}.
\newblock \bibinfo{journal}{\emph{Management Science}} \bibinfo{volume}{72}, \bibinfo{number}{1} (\bibinfo{year}{2025}), \bibinfo{pages}{242--264}.
\newblock
\href{https://doi.org/10.1287/mnsc.2022.03918}{doi:\nolinkurl{10.1287/mnsc.2022.03918}}


\bibitem[You et~al\mbox{.}(2022)]%
        {You2022AlgorithmicVersusHuman}
\bibfield{author}{\bibinfo{person}{Sangseok You}, \bibinfo{person}{Cathy~Liu Yang}, {and} \bibinfo{person}{Xitong Li}.} \bibinfo{year}{2022}\natexlab{}.
\newblock \showarticletitle{Algorithmic versus Human Advice: Does Presenting Prediction Performance Matter for Algorithm Appreciation?}
\newblock \bibinfo{journal}{\emph{Journal of Management Information Systems}} \bibinfo{volume}{39}, \bibinfo{number}{2} (\bibinfo{year}{2022}), \bibinfo{pages}{336--365}.
\newblock
\href{https://doi.org/10.1080/07421222.2022.2063553}{doi:\nolinkurl{10.1080/07421222.2022.2063553}}


\end{thebibliography}
\end{document}